%% file: main.tex
\documentclass[11pt]{article}

\usepackage[final]{acl}

\usepackage{times}
\usepackage{latexsym}
\usepackage[T1]{fontenc}
\usepackage[utf8]{inputenc}
\usepackage{microtype}
\usepackage{inconsolata}
\usepackage{graphicx}
\usepackage{subcaption}
\usepackage{booktabs}
\usepackage{amsmath}
\usepackage{placeins}

\title{Efficient Benchmarking in Production: A Study of an Evolving LLM Agent}

\author{Yining She\thanks{Work done while at Meta.}  \\
  Carnegie Mellon University\\
  \texttt{yiningsh@andrew.cmu.edu} \\\And
  Lei Lin \\
  Meta\\
  \texttt{llin22@meta.com} \\}

\begin{document}
\maketitle

\begin{abstract}

\input{sections/00_abstract}

\end{abstract}

\input{sections/01_introduction}

\input{sections/02_problem_setup}
\input{sections/03_methods}
\input{sections/04_experimental_setup}
\input{sections/05_results}
\input{sections/06_discussion}
\input{sections/07_related_work}

\input{sections/08_limitations}

\bibliography{main}

\appendix
\input{sections/10_appendix}

\end{document}

%% file: sections/00_abstract.tex
Production LLM agents are evaluated repeatedly as they evolve, but full agent benchmarks are costly to rerun. We study efficient recurring evaluation for a production analytics agent serving
tens of thousands of monthly active users
and report first-hand deployment experience. Using 574 historical runs of the production benchmark, split chronologically into calibration and held-out periods, we compare random sampling, historical caching, fixed representative subsets, and IRT-based adaptive testing. The results show that multidimensional 2PL adaptive testing achieves the best overall score fidelity: executing 200 questions, 38.5\% of a full run, yields 1.03 pp of MAE. 
We nevertheless deployed difficulty-stratified fixed subsets because of their operational simplicity, and show they transfer without recalibration to five other agent families and remain stable across calibration windows as short as one day. 
Drawing on this deployment experience, we report practical recommendations for recurring production-agent evaluation.

%% file: sections/01_introduction.tex
\section{Introduction}
\label{sec:introduction}

\begin{figure*}[t]
\vspace{-0.75em}
  \centering
  \includegraphics[width=0.9\textwidth]{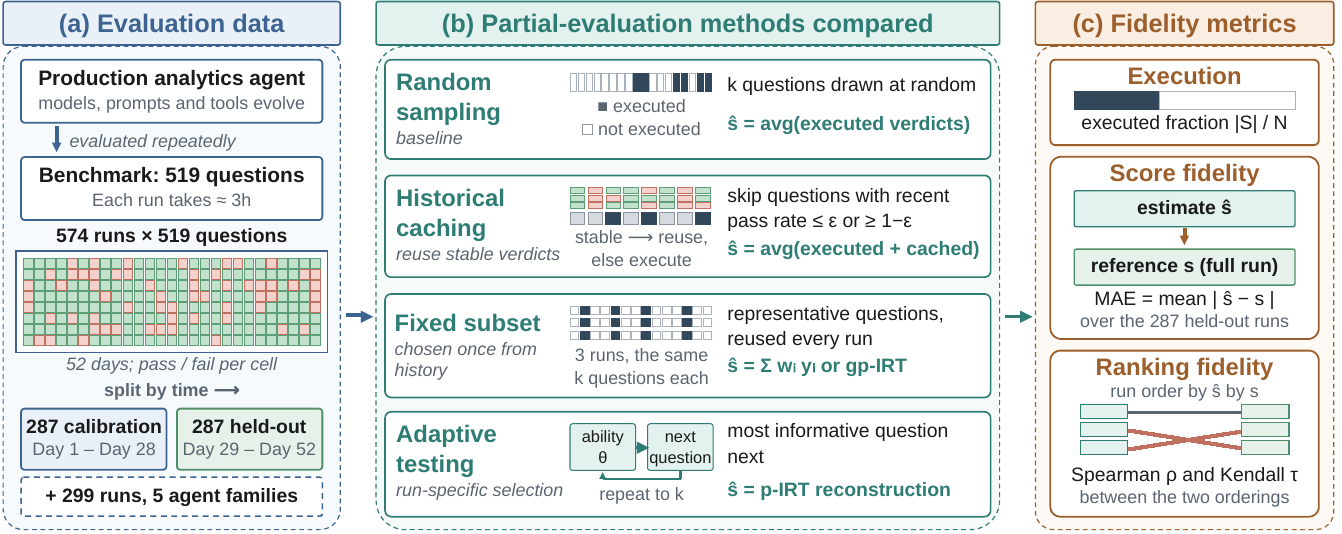}
  \caption{Overview of the study. \textbf{(a)} Historical evaluation runs of a
  production analytics agent, split by time into calibration and held-out test
  runs. \textbf{(b)} The four partial-evaluation methods compared: each executes
  a subset $S\subseteq\mathcal{Q}$ and reports an estimate $\hat{s}$ of the pass
  rate $s$ of a complete run. \textbf{(c)} Test on the 287 held-out runs:
  pass-rate MAE against the complete run (lower is better) and Spearman and
  Kendall rank correlation (higher is better).}
  \vspace{-0.75em}
  \label{fig:overview}
\end{figure*}

Production LLM agents require repeated evaluation as their models, prompts,
tools, and surrounding systems evolve. Each benchmark item may involve a
multi-step trajectory with repeated model calls, tool use, stateful interaction,
and task-specific grading; stochasticity may also require repeated trials
\citep{kapoor2026holistic,yao2024tau,jimenez2024swe}. 
Our work is motivated by this challenge in the development of a deployed analytics agent, which serves 
tens of thousands of monthly active users.
Its development and monitoring generate tens of thousands of evaluation runs.
Benchmark evaluation is a major component of this workload. Each run of a central 519-question benchmark 
takes approximately three hours. This
makes cost- and time-efficient
measurement important for continuous development.

Prior work reduces LLM evaluation cost through subset selection,
score reconstruction, and adaptive testing
\citep{vivek2024anchor,maiapolo2024tinybenchmarks,perlitz2024efficient,kipnis2025metabench,yuan2025beyond,truong2025reliable}. These methods are primarily calibrated on
responses of distinct models evaluated on public benchmarks.
Less is known about recurring evaluation of an evolving production agent, where
temporally ordered outcomes from earlier runs can inform later
evaluations.

We therefore ask how faithfully partial evaluation can stand in for the complete
benchmark throughout continuous agent development, along two axes:
\textbf{score fidelity}, the tradeoff between the number of questions executed and error
in the estimated full-benchmark score, and \textbf{ranking fidelity}, how
well rankings among evaluation runs are preserved by each method.

We compare random sampling with three uses of historical evaluation data:
caching reuses stable outcomes, fixed subsets execute representative
questions, and adaptive testing selects each next question based on responses in the current run \citep{maiapolo2024tinybenchmarks,truong2025reliable}. 
We use real historical data from our production agent: 574 benchmark runs collected over 52
days, split into 287 calibration and 287 held-out test runs (Fig.~\ref{fig:overview}).

Our results show that partial evaluation can closely recover full-benchmark
performance with substantially fewer executions. 
Multidimensional 2PL adaptive testing provides the strongest score fidelity:
executing 200 of 519 questions (38.5\% of a complete run) yields 1.03
percentage points (pp) of pass-rate MAE. 
Partial evaluation also
preserves rankings: at 300 questions, the best adaptive configuration reaches
0.992 Spearman correlation with full-benchmark rankings. Difficulty-stratified
fixed subsets give the strongest non-adaptive results, and we deployed them for
their operational simplicity. We further validate this choice through cross-agent transfer and calibration-window sensitivity.

Our contributions are as follows:\vspace{-0.5em}

\begin{itemize}\itemsep -0.3em
  \item We study recurring evaluation of an evolving production agent using 574
        historical benchmark runs and a temporally held-out protocol.
  \item We compare the score- and ranking-fidelity tradeoffs of random sampling,
        historical caching, fixed subsets, and adaptive testing.
  \item We analyze method design choices and validate difficulty-stratified
        fixed subsets through transfer to five agent families and sensitivity
        to calibration history.
  \item We derive practical recommendations from these results and our
        deployment experience in a production evaluation workflow.
\end{itemize}

%% file: sections/02_problem_setup.tex
\section{Problem Setup}
\label{sec:problem-setup}

\subsection{Recurring Production-Agent Evaluation}

As a production agent’s models, prompts, tools, and surrounding systems change,
developers need to repeatedly evaluate its configurations on a human-curated benchmark.
Each question receives a binary pass/fail verdict, which is aggregated into a
run-level pass rate used to compare configurations.

Let the complete benchmark contain $N$ questions,
$\mathcal{Q}=\{q_1,\ldots,q_N\}$. For evaluation run $r$, let
$y_{ri}\in\{0,1\}$ denote the verdict on question $q_i$, where one indicates a
pass. The complete vector of question-level verdicts is
\vspace{-0.3em}
\begin{equation}
  \mathbf{y}_r = (y_{r1},\ldots,y_{rN}),
\end{equation}\vspace{-0.5em}
and the complete-evaluation pass rate is
\begin{equation}
  s_r = \frac{1}{N}\sum_{i=1}^{N}y_{ri}.
\end{equation}
We treat $s_r$ as the reference score an efficient evaluation should recover.

\subsection{Partial Evaluation}

To reduce computational cost and latency, an efficient evaluation method $m$ executes a subset
$S_r^{(m)}\subseteq\mathcal{Q}$ and estimates the complete pass rate as
$\hat{s}_r^{(m)}$. Let $\mathcal{H}_r$ denote question-level verdicts available
from earlier runs. The estimator is
\begin{equation}
  \hat{s}_r^{(m)} =
  f_m\!\left(\{y_{ri}:q_i\in S_r^{(m)}\},\mathcal{H}_r\right),
\end{equation}
where $f_m$ may use current-run verdicts and history. Different methods may select different questions and compute their reported pass rates differently. Section~\ref{sec:methods} specifies each method’s selection rule and estimator.

We measure efficiency by the fraction of benchmark questions executed:
\begin{equation}
  \operatorname{ExecutionFraction}_r^{(m)}
  = |S_r^{(m)}|~/~{N}.
\end{equation}

\subsection{Evaluation Objectives}

Our first objective is \emph{score fidelity}: an efficient evaluation should
report a pass rate close to the complete evaluation. For a set $\mathcal{R}$ of
evaluation runs, we summarize score error using
\begin{equation}
  \operatorname{MAE}^{(m)} =
  \frac{1}{|\mathcal{R}|}
  \sum_{r\in\mathcal{R}}
  \left|\hat{s}_r^{(m)}-s_r\right|.
\end{equation}

Our second objective is \emph{ranking fidelity}. Developers use run rankings to
compare candidate changes and investigate regressions, so partial evaluation
should preserve the complete-evaluation ordering. We measure agreement between
$\{\hat{s}_r^{(m)}\}_{r\in\mathcal{R}}$ and $\{s_r\}_{r\in\mathcal{R}}$ using
Spearman correlation and Kendall’s Tau.

%% file: sections/03_methods.tex
\section{Methods}
\label{sec:methods}
We compare random sampling, historical caching, fixed subsets, and adaptive
testing. Each defines an executed set $S_r^{(m)}$ and an estimator of the
complete pass rate.

\subsection{Random Sampling}
\label{sec:method-random}

Random sampling draws $k$ questions uniformly without replacement and reports
their unweighted pass rate:
\begin{equation}
  \hat{s}_r^{\mathrm{rand}}
  = \frac{1}{k}\sum_{q_i\in S_r^{\mathrm{rand}}}y_{ri}.
\end{equation}

\subsection{Historical Outcome Caching}
\label{sec:method-cache}
Questions that consistently pass or fail across recent runs are less likely to change
their outcomes after a modest agent update.
Historical caching leverages this stability by reusing expected verdicts for selected questions instead of executing them again.
From the available history $\mathcal{H}_r$, we
select a look-back window $\mathcal{W}_r$ and compute the historical
pass rate of each question $q_i$:
\begin{equation}
  \bar{y}_{ri}
  = \frac{1}{|\mathcal{W}_r|}
    \sum_{u\in\mathcal{W}_r}y_{ui}.
\end{equation}
A question is cache eligible when
\begin{equation}\label{equation:cache-eligible}
  \min(\bar{y}_{ri},1-\bar{y}_{ri}) \leq \epsilon,
\end{equation}
where $\epsilon\in[0,0.5]$ controls the required stability. Eligible questions
receive their recent majority verdict,
$\tilde{y}_{ri}=\mathbf{1}[\bar{y}_{ri}\geq 0.5]$, while all others are executed.
If $K_r$ is the cached set, the method executes
$S_r=\mathcal{Q}\setminus K_r$ and reports
\begin{equation}
  \hat{s}_r^{\mathrm{cache}}
  = \frac{1}{N}\left(
      \sum_{q_i\in S_r}y_{ri}
      + \sum_{q_i\in K_r}\tilde{y}_{ri}
    \right).
\end{equation}

\subsection{Fixed-Subset Evaluation}
\label{sec:method-fixed}

Inspired by \citet{maiapolo2024tinybenchmarks},
fixed-subset evaluation uses historical outcomes to select a representative set
that is reused across later runs. We compare three selection strategies and two
score estimators to separate subset construction from score reconstruction.

\subsubsection{IRT Models}
Several selectors and estimators use item response theory (IRT) models fitted to
the historical response matrix. A multidimensional two-parameter logistic (2PL)
model assigns each run an ability vector $\boldsymbol{\theta}_r$, each question
a discrimination vector $\boldsymbol{\alpha}_i$, and each question a scalar
difficulty $\beta_i$. The model predicts the probability of a pass as
\begin{equation}
  p_{ri}
  = \Pr(y_{ri}=1\mid\boldsymbol{\theta}_r,
                  \boldsymbol{\alpha}_i,\beta_i)
  = \sigma(\boldsymbol{\alpha}_i^\top\boldsymbol{\theta}_r-\beta_i),
\end{equation}
where $\sigma$ is the logistic sigmoid. We estimate parameters from historical
verdicts by maximum a posteriori estimation with Gaussian priors. The Rasch, or
one-parameter logistic (1PL), model is the unidimensional special case with
$\alpha_i=1$:
\begin{equation}\label{equation:rasch}
  p_{ri}=\sigma(\theta_r-\beta_i).
\end{equation}

\subsubsection{Subset Selection}
We use difficulty stratification as a direct, interpretable construction that selects questions across the estimated difficulty range. For comparison, we also evaluate
historical-response and IRT-feature clustering, adapting the two clustering selectors of
\citet{maiapolo2024tinybenchmarks}.

\paragraph{Difficulty stratification.}
We fit the Rasch model (Eq.\ref{equation:rasch}) on the calibration runs, sort questions by difficulty $\beta_i$,
and divide them into $k$ approximately equal strata. From each
stratum, we select the question closest to its median difficulty, and weight it by the stratum's fraction of the
benchmark.

\paragraph{Historical-response clustering.}
We represent each question by a vector recording its pass/fail outcomes across
historical runs and apply K-means clustering with $k$ clusters. The question
closest to each cluster centroid is selected, and its weight is the fraction of
benchmark questions in that cluster.

\paragraph{IRT-feature clustering.}
We instead represent each question by its fitted 2PL item parameters,
$[\boldsymbol{\alpha}_i;\beta_i]$, and apply the same K-means selection and weighting.

\subsubsection{Score Estimation}
Following \citet{maiapolo2024tinybenchmarks}, we evaluate each selector with two
score estimators. The weighted subset estimator uses only the selected
questions, while generalized p-IRT estimator (gp-IRT) reconstructs the unobserved questions
using the fitted IRT model.

\paragraph{Weighted subset estimator.}
This estimator treats each selected question as the representative of its
group.
For fixed subset $S$, let $w_i$ be the cluster or stratum weight of question
$q_i$, with $\sum_{q_i\in S}w_i=1$. The weighted estimate is
\begin{equation}
  \hat{s}_r^{\mathrm{weighted}}
  = \sum_{q_i\in S}w_i y_{ri}.
\end{equation}

\paragraph{Generalized p-IRT estimator.}

The gp-IRT estimator combines the weighted subset estimate with an IRT-based
reconstruction of the entire benchmark, balancing sampling variation against
error from the fitted IRT model.
The IRT-based reconstruction is p-IRT. It first estimates the current run’s ability
$\hat{\boldsymbol{\theta}}_r$ from its responses in $S$, holding fitted item
parameters fixed, and then replaces each unobserved verdict with its predicted pass
probability:
\begin{equation}\label{equation:pirt}
  \hat{s}_r^{\mathrm{pIRT}}
  = \frac{1}{N}\left(
      \sum_{q_i\in S}y_{ri}
      + \sum_{q_i\notin S}p_i(\hat{\boldsymbol{\theta}}_r)
    \right).
\end{equation}
In our experiments, difficulty stratification uses Rasch parameters for selection and
reconstruction, while the clustering variants use the fitted multidimensional 2PL
model.

The gp-IRT estimator blends the weighted and p-IRT estimates:
\begin{equation}
  \hat{s}_r^{\mathrm{gpIRT}}
  = \lambda\hat{s}_r^{\mathrm{weighted}}
    +(1-\lambda)\hat{s}_r^{\mathrm{pIRT}},
\end{equation}
where the blending weight 
$\lambda$ is:
\begin{equation}
  \lambda =
  \frac{\hat{b}^{2}}
       {\hat{\sigma}^{2}/(4|S|)+\hat{b}^{2}}.
\end{equation}
The weight favors the direct estimate when estimated IRT reconstruction bias
$\hat{b}^2$ is large relative to weighted-estimator variance
$\hat{\sigma}^2/(4|S|)$. We estimate $\hat{\sigma}^2$ by averaging within-run
question-outcome variance over historical runs and $\hat{b}^2$ as squared IRT
reconstruction MAE on held-out historical runs. Following the default
anchor-estimator adjustment of \citet{maiapolo2024tinybenchmarks}, we divide
$\hat{\sigma}^2$ by four, equivalently halving the estimated standard deviation.

\subsection{IRT-Based Adaptive Testing}
\label{sec:method-adaptive}
Adaptive testing selects each next question using responses observed in the
current run. The intuition is to always pick the questions
that are most informative for the current evaluation. We adapt the Rasch/Fisher-information procedure of
\citet{truong2025reliable} and additionally evaluate a multidimensional 2PL extension with
D-optimal selection.

We first calibrate item difficulties $\{\beta_i\}_{i=1}^N$ from historical runs
and initialize a new run’s ability at $\hat{\theta}_r^{(0)}=0$. At step $t$, the
predicted pass probability for each unobserved question is
\begin{equation}
  p_i^{(t)}=\sigma(\hat{\theta}_r^{(t)}-\beta_i).
\end{equation}
Under the Rasch model, the Fisher information supplied by question $q_i$ is
\begin{equation}
  \mathcal{I}_i(\hat{\theta}_r^{(t)})
  =p_i^{(t)}(1-p_i^{(t)}).
\end{equation}
At each step, the method executes the unobserved question with maximum
information and re-estimates $\hat{\theta}_r$ from all observed verdicts,
stopping after $k$ questions.

The multidimensional 2PL variant estimates ability vector
$\hat{\boldsymbol{\theta}}_r$ and uses D-optimality to reduce uncertainty across
its coordinates. Given the accumulated Fisher information matrix $\mathbf{I}_t$,
it selects the question maximizing
$p_i^{(t)}(1-p_i^{(t)})\boldsymbol{\alpha}_i^\top
\mathbf{I}_t^{-1}\boldsymbol{\alpha}_i$, then updates the ability vector and
information matrix from all observed verdicts.

Because Fisher selection targets informative rather than representative
questions, the raw pass rate of the selected questions need not approximate the
complete-benchmark pass rate. We therefore estimate
the complete pass rate using the p-IRT reconstruction defined above (Eq.\ref{equation:pirt})

%% file: sections/04_experimental_setup.tex
\section{Experimental Setup}
\label{sec:experimental-setup}

\subsection{Production Evaluation Data}

The evaluated system is a production analytics agent designed to answer
natural-language questions about data. It combines language-model reasoning
with tools for retrieving and analyzing information, and is evaluated
repeatedly as its models, prompts, tools, and surrounding system change.

The benchmark consists of human-curated questions about data written by internal analysts; neither the questions nor the recorded evaluation outcomes contain user data or personal information. Each recorded outcome is a binary pass/fail verdict produced by an automated grader. We
retain evaluation runs with valid outcomes for at least 80\% of the benchmark
and compute each run's reference pass rate over its valid outcomes. 
This yields 287 calibration runs from Day 1 through Day 28, and 287 test runs from Day 29 through Day 52 (Figure~\ref{fig:overview}).

\subsection{Temporal Protocol and Method Configurations}

\paragraph{Random sampling.}
We evaluate $k\in\{10,\allowbreak20,\allowbreak30,\allowbreak50,\allowbreak75,\allowbreak100,\allowbreak150,\allowbreak200,\allowbreak250,\allowbreak300,\allowbreak350,\allowbreak400\}$, averaging
results over 20 random seeds.

\paragraph{Historical caching.}
Caching uses a rolling seven-day window of prior routine runs, requires at least
five historical trials per eligible question, and is rebuilt daily for reuse on
that day’s evaluations. We sweep $\epsilon$ from 0.00 to 0.30 in increments of
0.01, caching a question when
Eq.\ref{equation:cache-eligible} is satisfied.

\paragraph{Fixed-subset evaluation.}
We construct each subset once from all calibration runs at the same computational budgets $k$ used for random sampling and apply it unchanged to every test run. IRT models for selection and
p-IRT reconstruction are also fitted on all calibration runs. To estimate the
gp-IRT bias term $\hat{b}^2$, we split calibration runs in half, fit an auxiliary
IRT model on the first half, and measure reconstruction error on the second by
revealing half of each run’s available outcomes. For every multidimensional 2PL
fit, we select $d\in\{2,5,10,15\}$ by log likelihood on a fixed 15\% sample of
observed calibration cells.

\paragraph{Adaptive testing.}
We evaluate adaptive testing at the same $k$. Item parameters are fitted on all calibration runs and held fixed during
question selection for each test run. We compare Rasch adaptive testing with the multidimensional 2PL
variant to assess whether modeling multiple latent abilities improves the
accuracy--execution tradeoff.

%% file: sections/05_results.tex
\section{Results}
\label{sec:results}

\subsection{Score Fidelity}
\label{sec:results-score}

\begin{figure}[t]
  \centering
  \includegraphics[width=0.9\columnwidth]{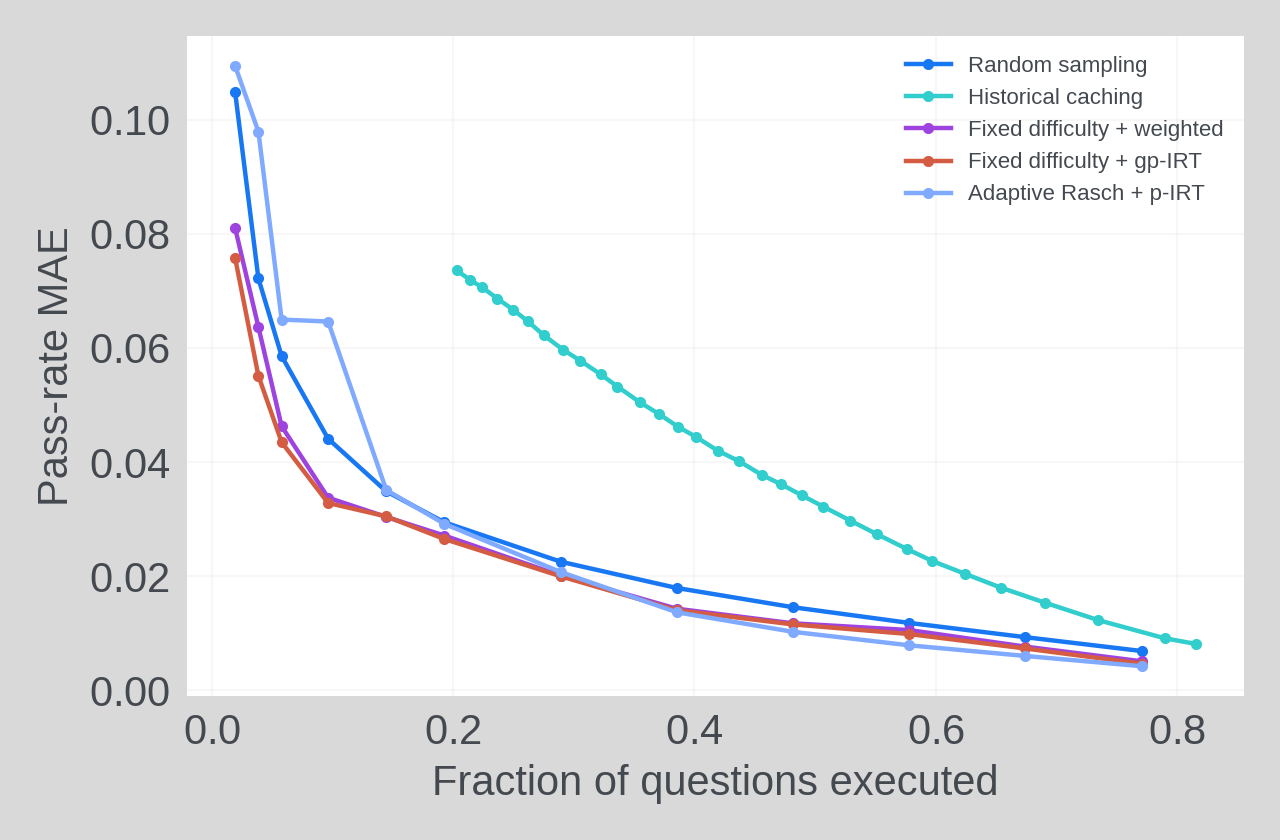}
  \caption{Pass-rate MAE v.s. execution fraction.
     Fixed-budget methods use $k/519$. Caching points report mean execution
     fractions across cache thresholds.}
  \label{fig:main-mae}
  \vspace{-0.75em}
\end{figure}

Figure~\ref{fig:main-mae} compares historical caching,
difficulty-stratified fixed subsets with weighted and gp-IRT estimation, and
Rasch adaptive testing on 287 held-out runs\footnote{Appendix report
complete values in tables.}.
We examine the remaining design choices within each method family in
Section~\ref{sec:results-design}.

Among the methods evaluated at fixed computational budgets $k$, difficulty-stratified subsets lead at small budgets: at $k{=}100$ (19.3\% of the benchmark), gp-IRT obtains
2.65pp MAE, compared with 2.94 for random sampling and 2.92
for adaptive testing. Adaptive testing leads from $k{=}200$ (38.5\%), where its MAE is 1.37
pp versus 1.40 for fixed-subset gp-IRT and 1.79 for random sampling. At $k{=}300$ (57.8\%),
the respective errors are 0.79, 0.99, and 1.18 pp. Weighted and gp-IRT subset
estimates are similar across these budgets.

Caching determines execution through its eligibility threshold, yielding mean
execution fractions from 20.3\% to 81.6\% and MAE from 7.36 to 0.81 pp. More
aggressive caching reduces execution but increases error, and its curve lies
above the other methods over their shared range. For example, caching obtains
1.54 pp MAE at 69.1\% execution, while adaptive testing, fixed-subset gp-IRT,
and random sampling obtain 0.60, 0.73, and 0.93 pp at $k{=}350$ (67.4\%).

\paragraph{Takeaway.}
Difficulty-stratified fixed subsets perform best at small budgets, while adaptive testing achieves the lowest MAE from $k{=}200$ onward. Historical caching is less accurate than the other methods at comparable execution levels.

\subsection{Ranking Fidelity}
\label{sec:results-ranking}

\begin{figure*}[t]
  \centering
  \begin{minipage}[t]{0.45\textwidth}
    \centering
    \includegraphics[width=\linewidth]{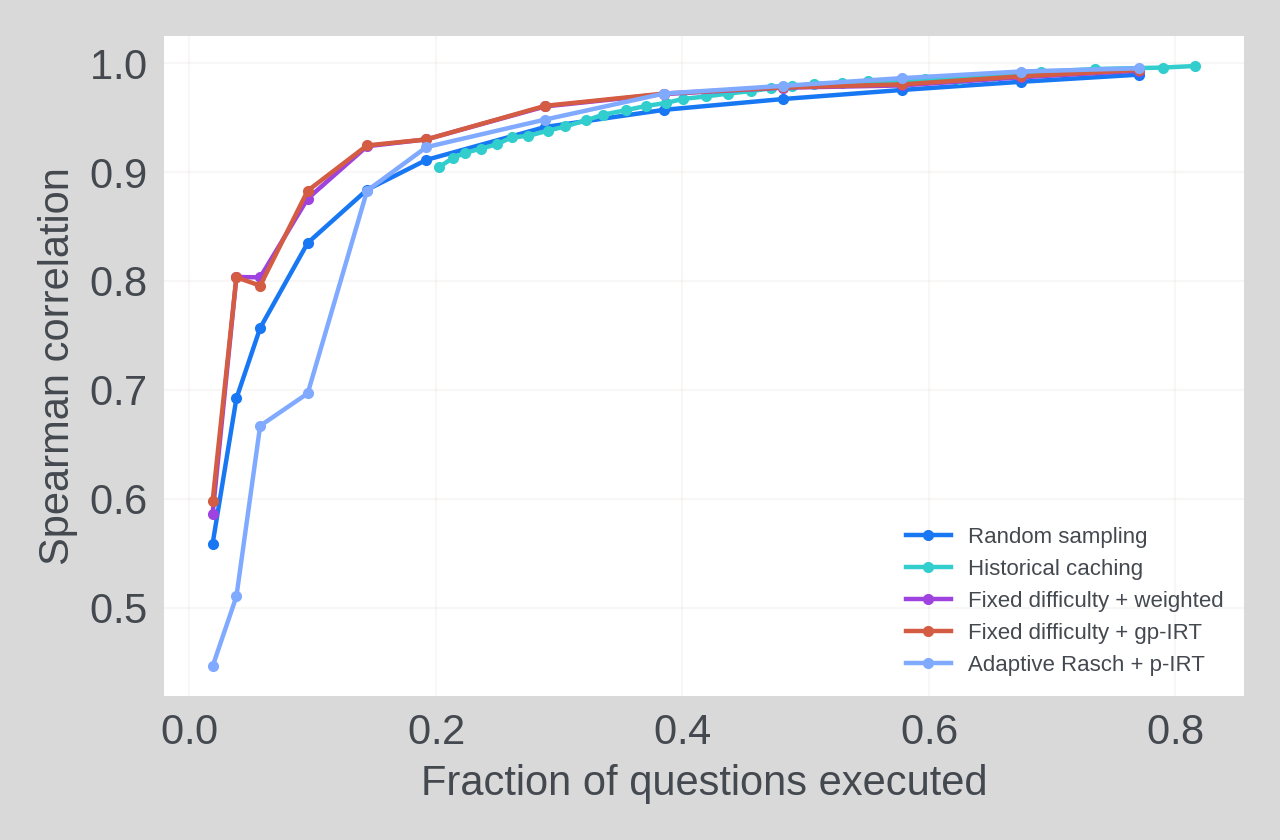}
    \small (a) Spearman correlation
  \end{minipage}\hfill
  \begin{minipage}[t]{0.45\textwidth}
    \centering
    \includegraphics[width=\linewidth]{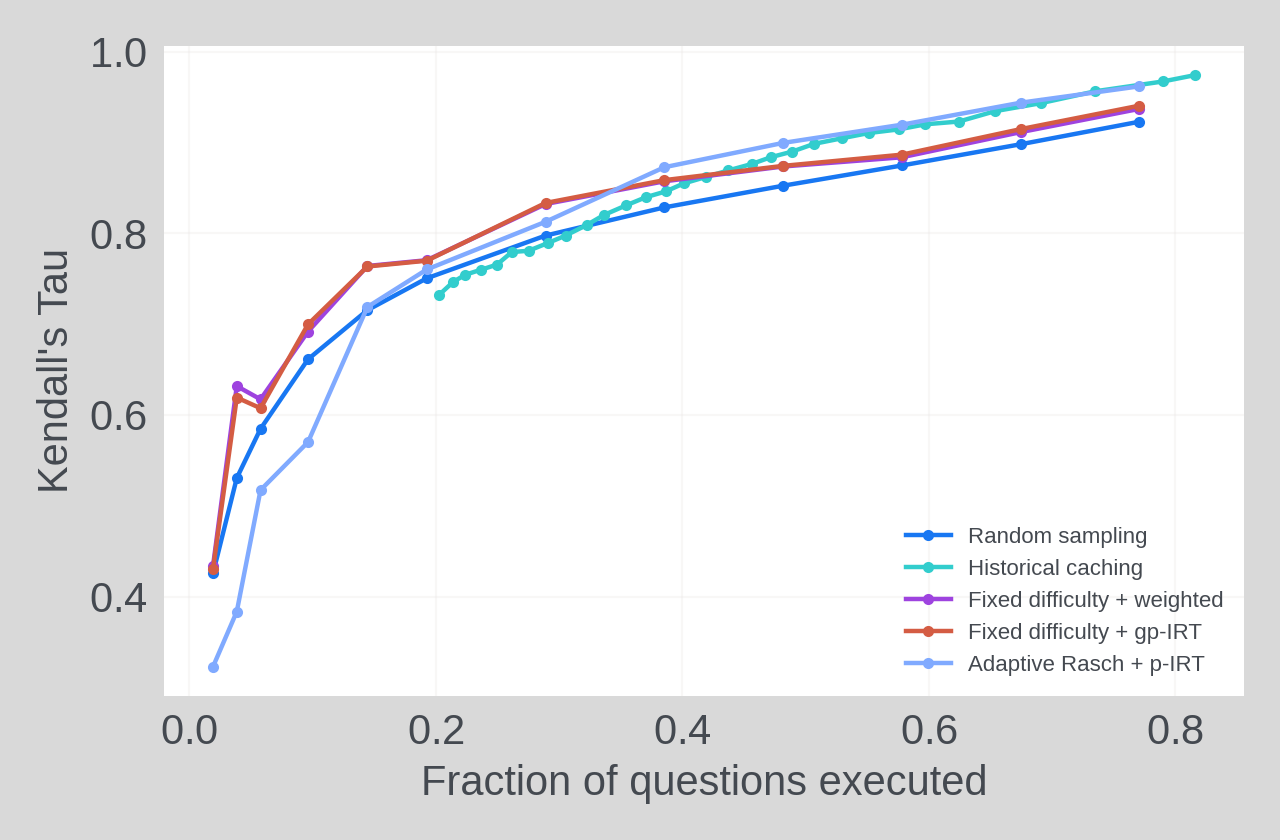}
    \small (b) Kendall's Tau
  \end{minipage}
  \vspace{-0.75em}
  \caption{Ranking fidelity v.s. execution fraction on 287 held-out runs.
  Higher is better.}
  \label{fig:main-ranking}
  \vspace{-0.75em}
\end{figure*}

Fig.~\ref{fig:main-ranking} reports ranking fidelity for the same
configurations. Both correlations increase with the computational budget, but the differences among methods are more pronounced for Kendall's Tau.
At $k{=}100$ (19.3\%), fixed-subset gp-IRT obtains 0.930 Spearman and 0.770 Kendall,
compared with 0.923\&0.760 for adaptive testing and 0.911\&0.751 for random sampling. 

Adaptive testing becomes stronger as the computational budget increases.
At $k{=}200$ (38.5\%), adaptive testing and fixed-subset gp-IRT tie in Spearman
(0.972), while adaptive testing has higher Kendall (0.873 v.s. 0.859). At
$k{=}300$ (57.8\%), adaptive testing reaches 0.986 Spearman and 0.920 Kendall,
compared with 0.981\&0.887 for fixed-subset gp-IRT and 0.975\&0.875 for
random sampling across budgets.

Caching preserves rankings much better than its score MAE suggests. At 20.3\%
execution it obtains 0.905 Spearman and 0.733 Kendall. At 79.0\%, it reaches
0.996 and 0.967, compared with 0.995 and 0.962 for adaptive testing at $k{=}400$
(77.1\%). 

\paragraph{Takeaway.}
Difficulty-stratified subsets give the strongest ranking fidelity at small
budgets, while adaptive testing leads at larger budgets. Historical caching
preserves run rankings substantially better than its score-estimation error
alone would suggest.

\subsection{Design Choices within Method Families}
\label{sec:results-design}

The score and ranking fidelity comparisons in
Sections~\ref{sec:results-score} and~\ref{sec:results-ranking} yield
similar method orderings. We therefore focus on MAE in
the remaining analyses and report rank
correlations in the appendix.

\subsubsection{Fixed-Subset Selection and Estimation}
\begin{figure}[t]
  \centering
  \includegraphics[width=0.9\columnwidth]{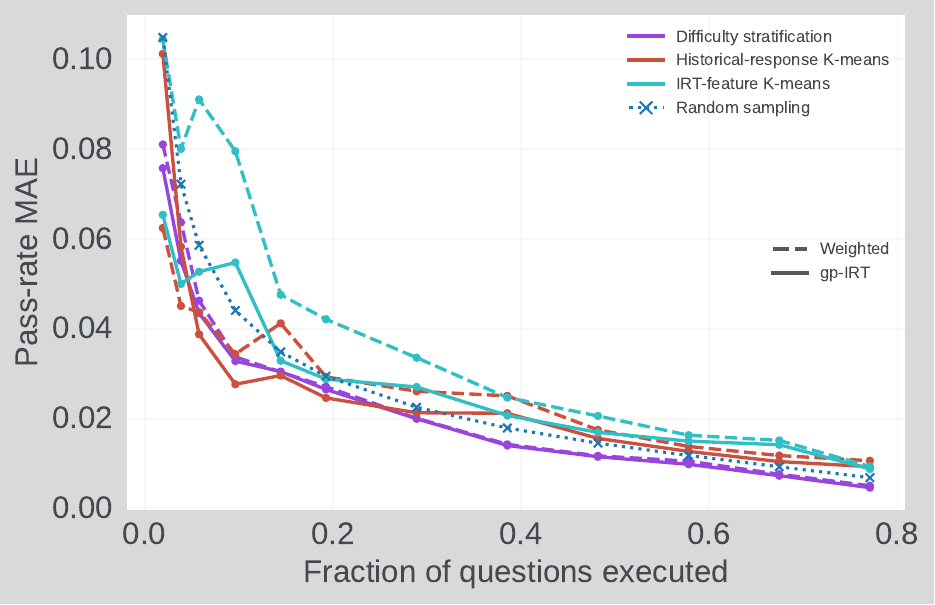}
  \caption{Fixed-subset MAE on 287 held-out runs.}
  \label{fig:fixed-subset-mae}
  \vspace{-0.75em}
\end{figure}

Figure~\ref{fig:fixed-subset-mae} separates the effect of subset selection from
score estimation.
Difficulty stratification has the lowest MAE across most budgets and leads under
both estimators from $k{=}150$ (28.9\%). With gp-IRT, historical-response K-means
leads from $k{=}30$ (5.8\%) through $k{=}100$ (19.3\%) and generally outperforms
IRT-feature K-means. From $k{=}200$ (38.5\%), both clustering selectors perform
worse than random sampling under either estimator, while difficulty
stratification remains better.

gp-IRT benefits the clustering selectors more than difficulty stratification,
whose weighted and gp-IRT estimates are consistently similar. Ranking correlation results largely mirror MAE (Fig.~\ref{fig:fixed-subset-spearman}\&\ref{fig:fixed-subset-kendall}).

The weak performance of IRT-feature K-means contrasts with \citet{maiapolo2024tinybenchmarks}, which found IRT-based selection consistently effective. Two differences between the experimental settings may explain the conflicts.
First, our calibration data of recurring configurations of one production agent may contain fewer
distinct response patterns than diverse LLM populations used by \citet{maiapolo2024tinybenchmarks}, making multidimensional item parameters less reliable and therefore clusters less representative.
Difficulty stratification requires only a scalar difficulty ordering and directly preserves coverage across that range.
Second, our subsets are much larger fractions of the benchmark: $k{=}100$ (19.3\%) versus 100 of roughly 14,000 MMLU questions in \citet{maiapolo2024tinybenchmarks}. 
Random sampling becomes more accurate at these larger sampling fractions, leaving less scope for clustering to improve upon it.

\paragraph{Takeaway.}
Difficulty stratification performs best overall, while the more complex
selectors provide no consistent gain and fall below random sampling at moderate computational 
budgets.

\subsubsection{Adaptive IRT Model}

\begin{figure}[t]
  \centering
  \includegraphics[width=0.9\columnwidth]{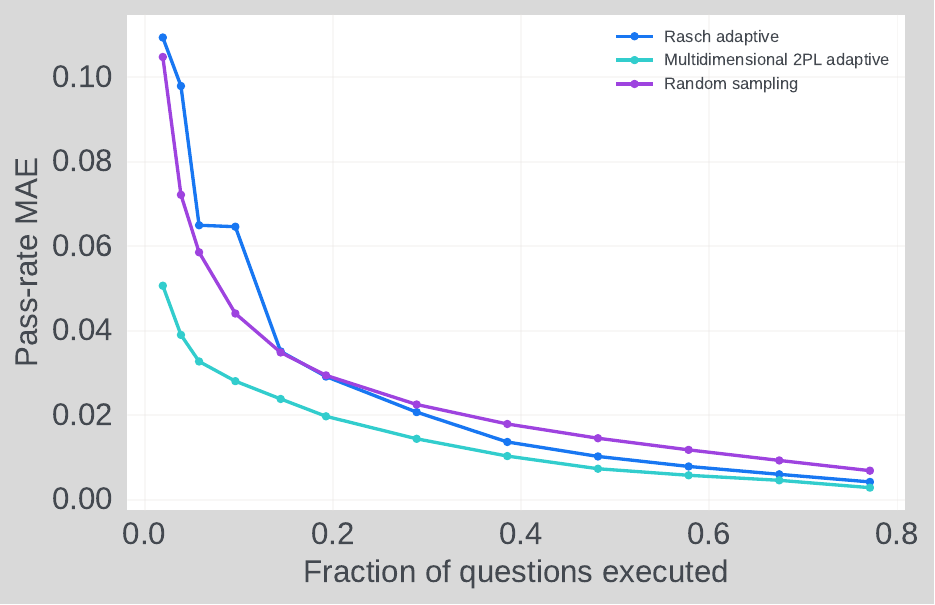}
  \caption{Adaptive Testing MAE on the held-out runs.}
  \label{fig:adaptive-model-mae}
  \vspace{-0.75em}
\end{figure}

Figure~\ref{fig:adaptive-model-mae} compares Rasch and multidimensional 2PL
adaptive testing.
Multidimensional 2PL adaptive testing has lower MAE than Rasch and random
sampling at every $k$, with its largest advantage at $k{=}100$ (19.3\%):
1.97 pp versus 2.92 and 2.94 pp, respectively. Its advantage narrows as execution increases. Spearman and Kendall show the same ordering (Figure~\ref{fig:adaptive-model-ranking}).

Rasch represents each run on one ability axis, whereas multidimensional 2PL
allows question outcomes to depend on different ability coordinates through
their discrimination vectors. D-optimal selection can therefore choose questions that inform ability coordinates not well covered by the questions already executed. This benefit is
largest at small computational budgets. As execution increases, p-IRT
replaces more predictions with observations, reducing the models' difference.

\paragraph{Takeaway.}
A more expressive IRT model can substantially improve adaptive testing.

\section{Practical Validation of Fixed-Subset}
\label{sec:fixed-validation}

The preceding comparisons show that difficulty-stratified fixed subsets provide
strong score and ranking fidelity. Although multidimensional adaptive testing
achieves lower error, difficulty stratification is operationally simpler: the
subset is constructed once and requires neither sequential selection nor
run-specific ability updates. Executing the same questions in every run also
supports direct question-level comparisons and diagnosis as the production
agent evolves. These properties motivated its use in our production workflow.

We therefore validate two properties important for recurring evaluation on difficulty-stratified fixed subset only:
whether a subset calibrated on one agent transfers to other agent families, and
how its fidelity changes with the calibration window.

\subsection{Cross-Agent Transfer}
\label{sec:cross-agent-transfer}

We apply the difficulty-stratified subsets and Rasch and gp-IRT parameters
calibrated on the original agent, without recalibration, to 299 runs from five
additional agent families. These distinct systems differ in their agent
frameworks, execution harnesses, tool sets, and model configurations. No
outcomes from these families are used to select questions or fit the estimators. We compare against uniform random sampling over the same computational budgets, averaged over 20 seeds.

\begin{figure}[t]
  \centering
  \includegraphics[width=0.9\columnwidth]{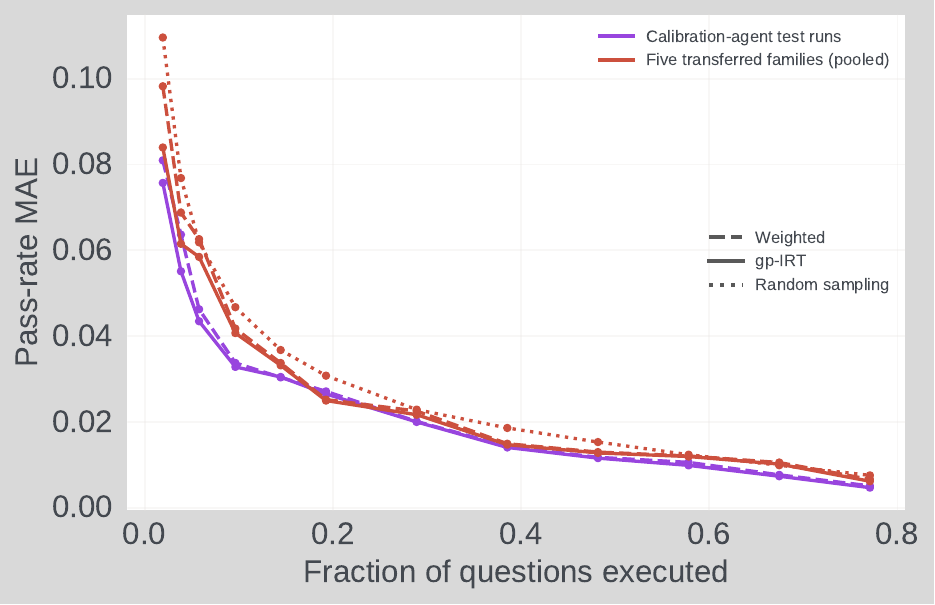}
  \caption{MAE on 287 original-agent and 299 transferred runs from five
  families. The subset and estimators use only original-agent calibration data.}
  \label{fig:cross-agent-transfer}
  \vspace{-0.75em}
\end{figure}

On the pooled transferred runs, gp-IRT has lower MAE than random sampling at 11
of 12 budgets (Figure~\ref{fig:cross-agent-transfer}). At $k{=}200$ (38.5\%),
random sampling, weighted estimation, and gp-IRT obtain 1.86, 1.49, and 1.47 pp
MAE, respectively; gp-IRT also remains close to its 1.40 pp MAE on the original
287 held-out runs. Both fixed-subset estimators also exceed random sampling in
Spearman and Kendall correlation at every budget
(Figure~\ref{fig:cross-agent-transfer-ranking}).

The advantage is not uniform across every family and computational budget. Nevertheless, gp-IRT has lower MAE than random sampling for a majority of budgets within each family. The appendix reports complete pooled and family-level results.

This transfer indicates that the IRT model captures intrinsic aspects of question difficulty that generalize across agent systems, even when calibrated on a single agent.

\paragraph{Takeaway.}
Without recalibration, the difficulty-stratified subset outperforms random
sampling on the pooled transferred runs at 11 of 12 budgets and remains close
to its original-agent fidelity, although gains vary across families.

\subsection{Calibration-Window Sensitivity}
\label{sec:calibration-window}

We next examine how the amount and time span of calibration data affect the
fixed subset. 
We fix the calibration end day at Day 28 and advance its start day in three-day increments from Day 1 to Day 28, 
yielding ten nested windows containing 287 to 14 runs (Table~\ref{tab:calibration-window-complete}).
For each window, we refit Rasch
and gp-IRT, reconstruct the difficulty-stratified subsets for
$k\in\{100,200,300,400\}$, and evaluate them on the same 287 held-out runs.

\begin{figure}[t]
  \centering
  \includegraphics[width=0.9\columnwidth]{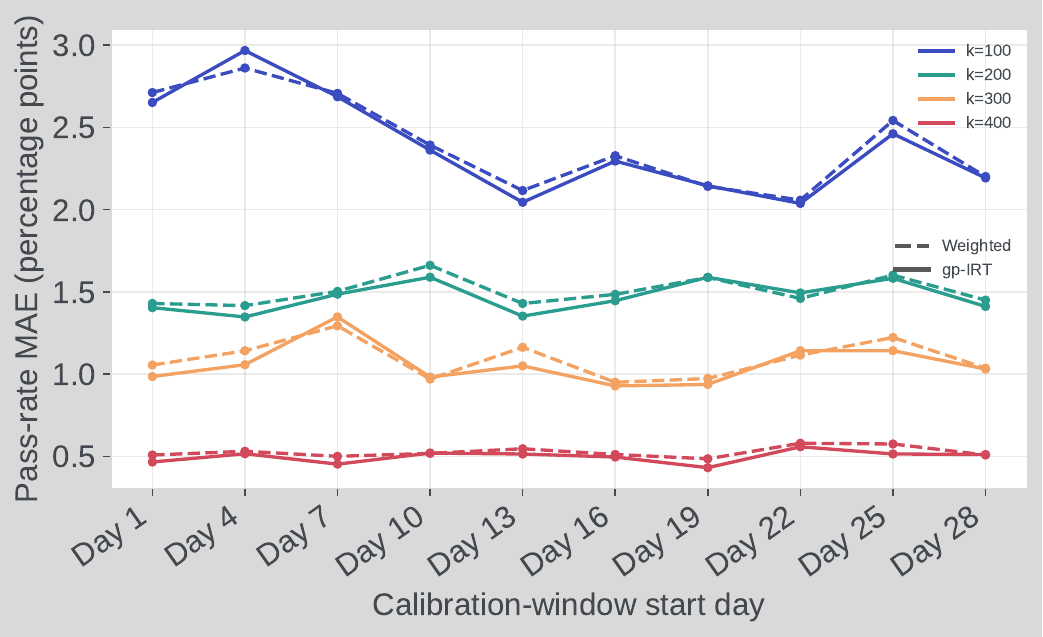}
  \caption{MAE across different calibration windows on 287 held-out runs.}
  \label{fig:calibration-window-mae}
  \vspace{-0.75em}
\end{figure}

MAE does not increase monotonically as the calibration window shortens
(Fig.~\ref{fig:calibration-window-mae}). Variation is greatest at $k{=}100$. At $k{=}200$, gp-IRT MAE ranges from 1.35-1.59 pp: the
one-day window (14 runs) obtains 1.41 pp, compared with 1.40 pp for the full
four-week window. Weighted and gp-IRT estimates remain similar, while Spearman
and Kendall show the same non-monotonic pattern
(Fig.~\ref{fig:calibration-window-spearman}\&\ref{fig:calibration-window-kendall}).

This stability may reflect the relatively mature development stage of the agent
and limited day-to-day change during the study. Recent runs may therefore
suffice to estimate the difficulty ordering used for stratification and may
better match the immediately following test period, while older runs add limited new information.

\paragraph{Takeaway.}
Fixed-subset fidelity is stable across the studied calibration windows,
with no consistent benefit from retaining the full 4-week history.

%% file: sections/06_discussion.tex
\section{Practical Recommendations}
\label{sec:discussion}

Production deployment requires considering
operational properties alongside statistical fidelity. Historical caching
retains the full question set, but requires recent outcomes and monitoring for stale verdicts. Fixed subsets provide predictable
execution and common questions across runs. Adaptive testing can
tailor selection to each run, but requires sequential orchestration,
run-specific state, and repeated ability updates. These operational differences
make the appropriate method dependent on the surrounding evaluation workflow.

For our production workflow, we deployed difficulty-stratified fixed subsets with $k\in\{100,200,300,400\}$, allowing users to choose an execution--fidelity tradeoff. 
Fixed subsets reveal the workload
before execution and preserve question-level comparisons for diagnosing
behavioral changes. They also avoid additional sequential serving logic in an evaluation
system designed to execute questions in parallel.
The subset transferred without recalibration to five additional agent families
and showed no monotonic loss of fidelity across the calibration windows. These
results support reuse across the systems and period studied.

As the agent evolves, we recommend recalibrating after material changes to its
models, prompts, tools, or execution system, or after sustained shifts in
question-level outcomes. Developers should also periodically run the full
benchmark and compare its pass rate with the fixed-subset estimate to measure
the error introduced during ongoing monitoring.

%% file: sections/07_related_work.tex
\section{Related Work}
\label{sec:related-work}

\paragraph{Efficient Evaluation.}

Prior work reduces evaluation cost by estimating complete-benchmark results
from item subsets. Fixed-form methods use cross-model correlations, IRT, clustering, or benchmark-level optimization
\citep{vivek2024anchor,maiapolo2024tinybenchmarks,perlitz2024efficient,kipnis2025metabench,fogliato-etal-2024-precise}.
Other methods tailor subsets to the target model, optimize ranking preservation,
or jointly select observed items and predict unobserved outcomes
\citep{yuan2025beyond,saranathan2025sublime,li2025active}. Adaptive evaluation,
building on computerized adaptive testing \citep{lord1980applications}, selects
questions by their expected information about the current system
\citep{truong2025reliable}. These methods are typically calibrated on response
matrices of distinct language models. We instead compare historical caching,
reusable fixed subsets, and adaptive testing on temporally ordered evaluations
of one evolving production agent.

\paragraph{Evaluating LLM Agents}
Agent evaluation is especially expensive because each item may require a full
trajectory involving repository interaction, browsing, tool calls, or
multi-turn interaction
\citep{jimenez2024swe,mialon2024gaia,yao2024tau}. 
Such work develops benchmarks and infrastructure
for comparing agents, while our setting concerns recurring evaluations generated as one agent evolves.

\paragraph{Continuous Evaluation of Evolving Systems}
Our setting also relates to regression-test minimization, selection, and
prioritization for evolving software
\citep{rothermel1997safe,yoo2012regression}. Continuous-integration methods use
code changes and historical executions to allocate testing resources
\citep{kim2002history,elbaum2014techniques,machalica2019predictive}, while
non-deterministic agent outcomes resemble flaky tests
\citep{luo2014empirical}. Conventional regression testing seeks faults using
signals such as coverage or change impact, but changes to an LLM agent do not
directly identify affected benchmark questions. We therefore use historical
question-level outcomes and a partial current run to estimate the complete
benchmark score and run ranking.

%% file: sections/08_limitations.tex
\section*{Limitations}
\label{sec:limitations}

Our data come from one organization and one production benchmark. Although the
difficulty-stratified subset transferred to five additional agent families,
this evidence covers only fixed subsets and does not establish generalization
to other benchmarks, historical caching, or adaptive testing. Future work should extend the analysis using data from a more diverse range of agents.

We evaluate pass-rate MAE and aggregate ranking fidelity using Spearman and
Kendall correlations, which do not directly measure regression detection or
release-gate decisions at operational thresholds. Future work should add threshold-based analyses that measure missed regressions, false alarms, and agreement with decisions based on the full benchmark.

We do not release the benchmark questions, the question-level outcome matrix, or the implementation. The methods we compare are fully specified in Section 3 and are reproducible on any benchmark with recorded per-item outcomes.

%% file: sections/10_appendix.tex
\section{Additional Results}
\label{sec:appendix-results}

This appendix reports the complete numerical results underlying the paper's
plots. The guide below follows the order of the corresponding main-text
analyses and points to the relevant figures and tables. In matched-budget
method comparisons, boldface marks the best displayed value and underlining
marks the second-best; table panel headings state the metric direction.

\subsection{Guide to the Complete Results}
\label{sec:appendix-guide}

\paragraph{Headline method comparison.}
The exact values behind the principal accuracy--efficiency
comparison is in Table~\ref{tab:main-fixed-budget-complete}; it reports
executed questions, execution fraction, pass-rate MAE, Spearman correlation,
and Kendall's $\tau$ for every method and operating point in
Figures~\ref{fig:main-mae} and~\ref{fig:main-ranking}.
Table~\ref{tab:cache-complete} gives the
corresponding threshold-level results for historical caching. Together, these
tables provide the complete numerical record for the two headline plots.

\begin{figure*}[t]
    \centering
    \includegraphics[width=0.82\textwidth]{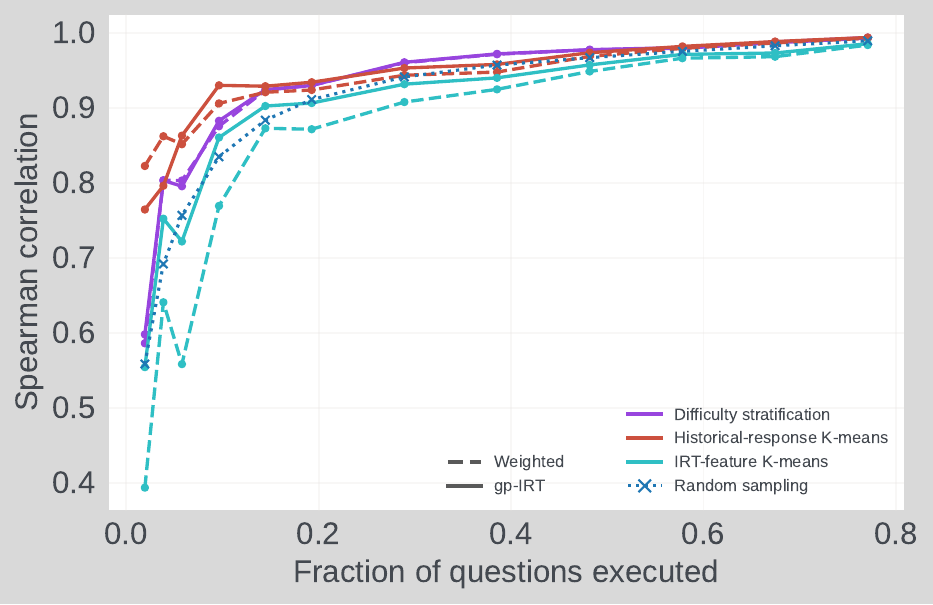}
    \caption{Spearman ranking fidelity for the fixed-subset
    selection-by-estimation comparison. Higher values are better. All methods
    use the same 287 held-out evaluation runs and the same question budgets;
    random sampling is averaged over 20 seeds.}
    \label{fig:fixed-subset-spearman}
\end{figure*}

\begin{figure*}[t]
    \centering
    \includegraphics[width=0.82\textwidth]{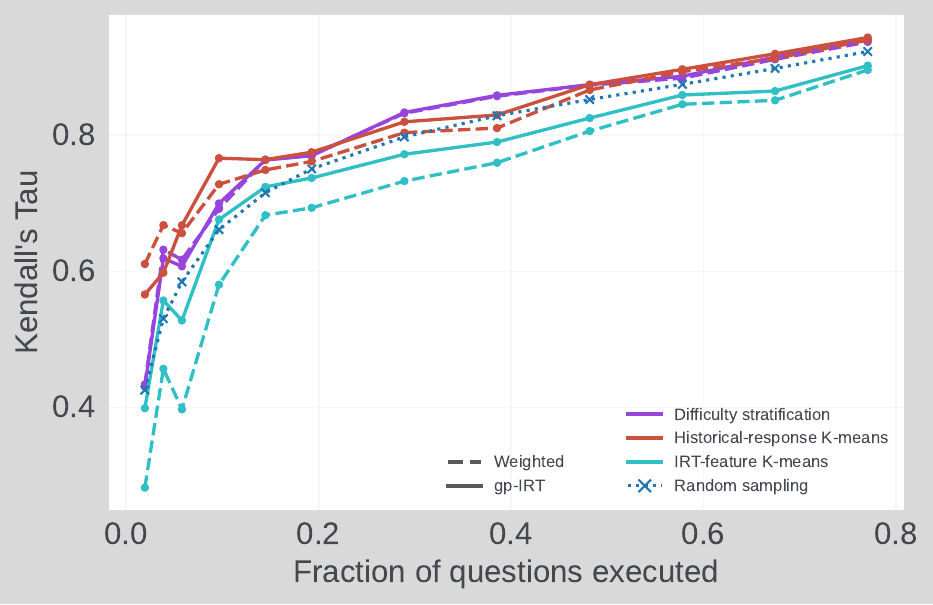}
    \caption{Kendall's $\tau$ ranking fidelity for the fixed-subset
    selection-by-estimation comparison. Higher values are better. The test
    population, budgets, and random-sampling seeds match
    Figure~\ref{fig:fixed-subset-spearman}.}
    \label{fig:fixed-subset-kendall}
\end{figure*}

\paragraph{Fixed-subset design results.}
Figures~\ref{fig:fixed-subset-spearman} and
\ref{fig:fixed-subset-kendall} give the full ranking-fidelity curves for the
selection-by-estimation comparison whose MAE results appear in
Figure~\ref{fig:fixed-subset-mae}. Table~\ref{tab:fixed-design-complete}
then collects all three metrics for every selector, estimator, and budget. 

\begin{figure*}[t]
    \centering
    \begin{subfigure}[t]{0.49\textwidth}
        \centering
        \includegraphics[width=\linewidth]{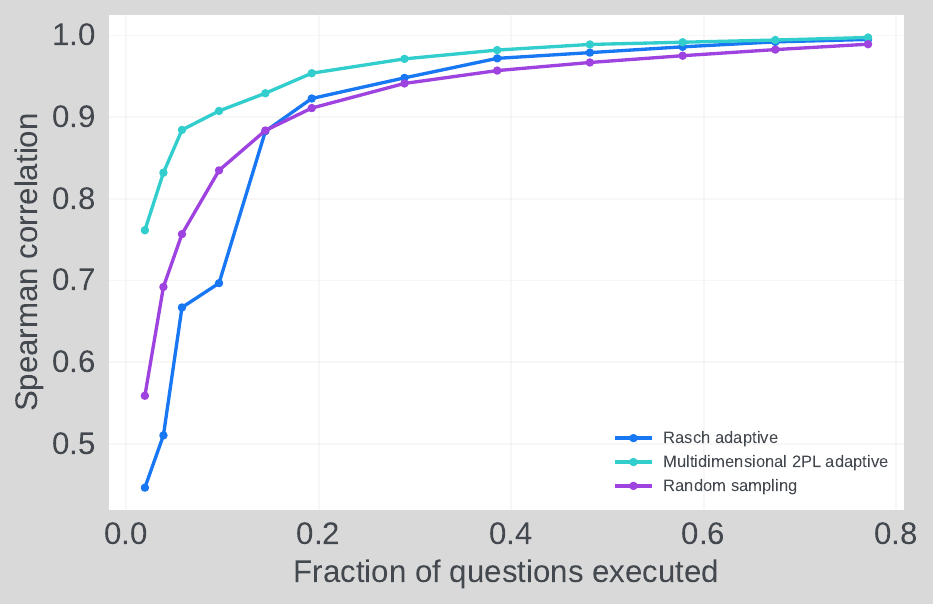}
        \caption{Spearman correlation.}
        \label{fig:adaptive-model-spearman}
    \end{subfigure}
    \hfill
    \begin{subfigure}[t]{0.49\textwidth}
        \centering
        \includegraphics[width=\linewidth]{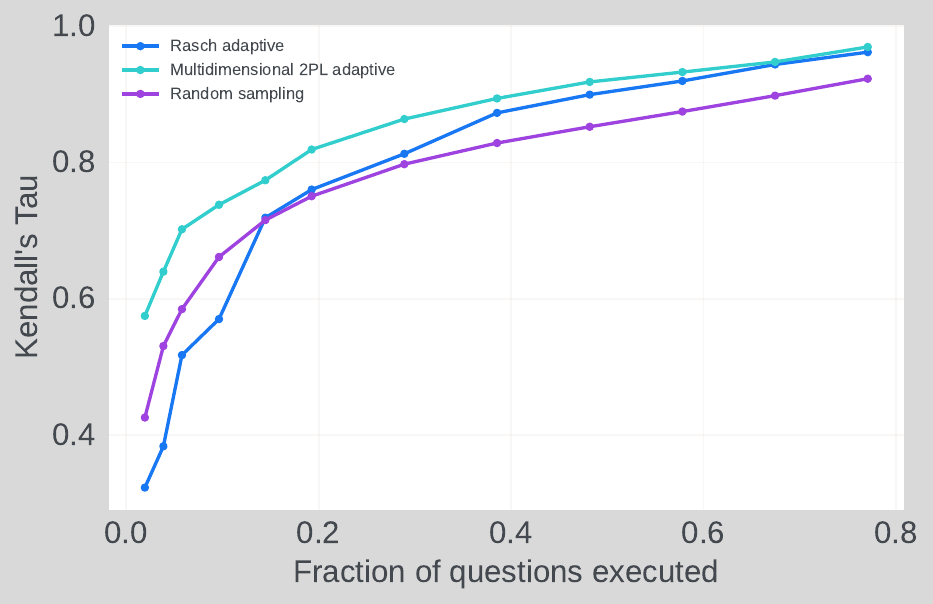}
        \caption{Kendall's $\tau$.}
        \label{fig:adaptive-model-kendall}
    \end{subfigure}
    \caption{Ranking fidelity for the adaptive-model comparison. Higher values
    are better. Both panels use the same 287 held-out evaluation runs and
    question budgets as Figure~\ref{fig:adaptive-model-mae}; random sampling is
    averaged over 20 seeds.}
    \label{fig:adaptive-model-ranking}
\end{figure*}

\paragraph{Adaptive-model design results.}
Figure~\ref{fig:adaptive-model-ranking} supplies the Spearman and Kendall views
corresponding to the MAE comparison in Figure~\ref{fig:adaptive-model-mae}.
Table~\ref{tab:adaptive-model-complete} reports the exact MAE and rank correlations
for Rasch adaptive testing, multidimensional 2PL adaptive testing, and random
sampling at every evaluated budget.

\begin{figure*}[t]
    \centering
    \begin{subfigure}[t]{0.49\textwidth}
        \centering
        \includegraphics[width=\linewidth]{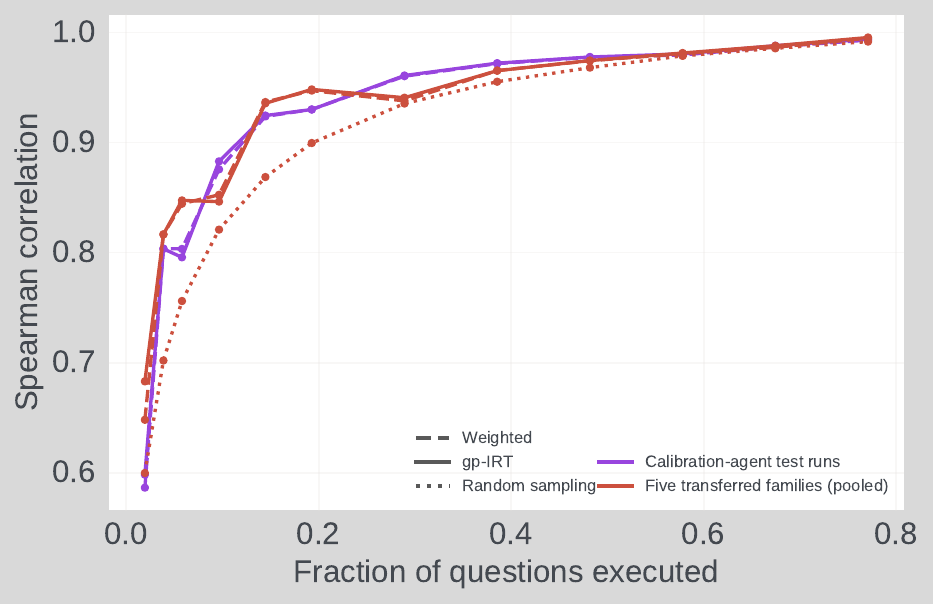}
        \caption{Spearman correlation.}
        \label{fig:cross-agent-transfer-spearman}
    \end{subfigure}
    \hfill
    \begin{subfigure}[t]{0.49\textwidth}
        \centering
        \includegraphics[width=\linewidth]{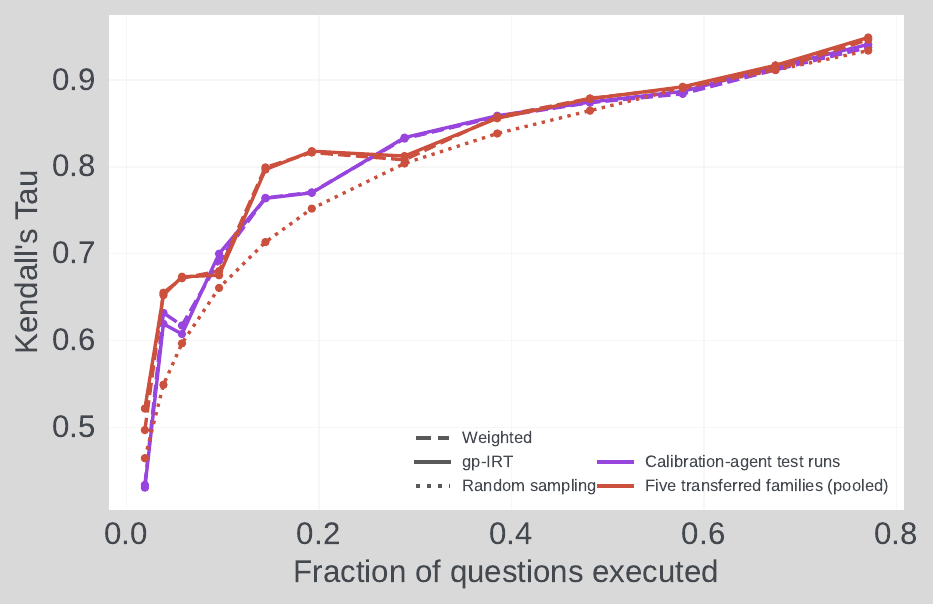}
        \caption{Kendall's $\tau$.}
        \label{fig:cross-agent-transfer-kendall}
    \end{subfigure}
    \caption{Pooled cross-agent ranking fidelity for a
    difficulty-stratified gp-IRT subset calibrated on the focal family and
    transferred unchanged to 299 runs from five held-out families. Higher
    values are better. Random sampling is averaged over 20 seeds.}
    \label{fig:cross-agent-transfer-ranking}
\end{figure*}

\begin{figure*}[t]
    \centering
    \includegraphics[width=0.82\textwidth]{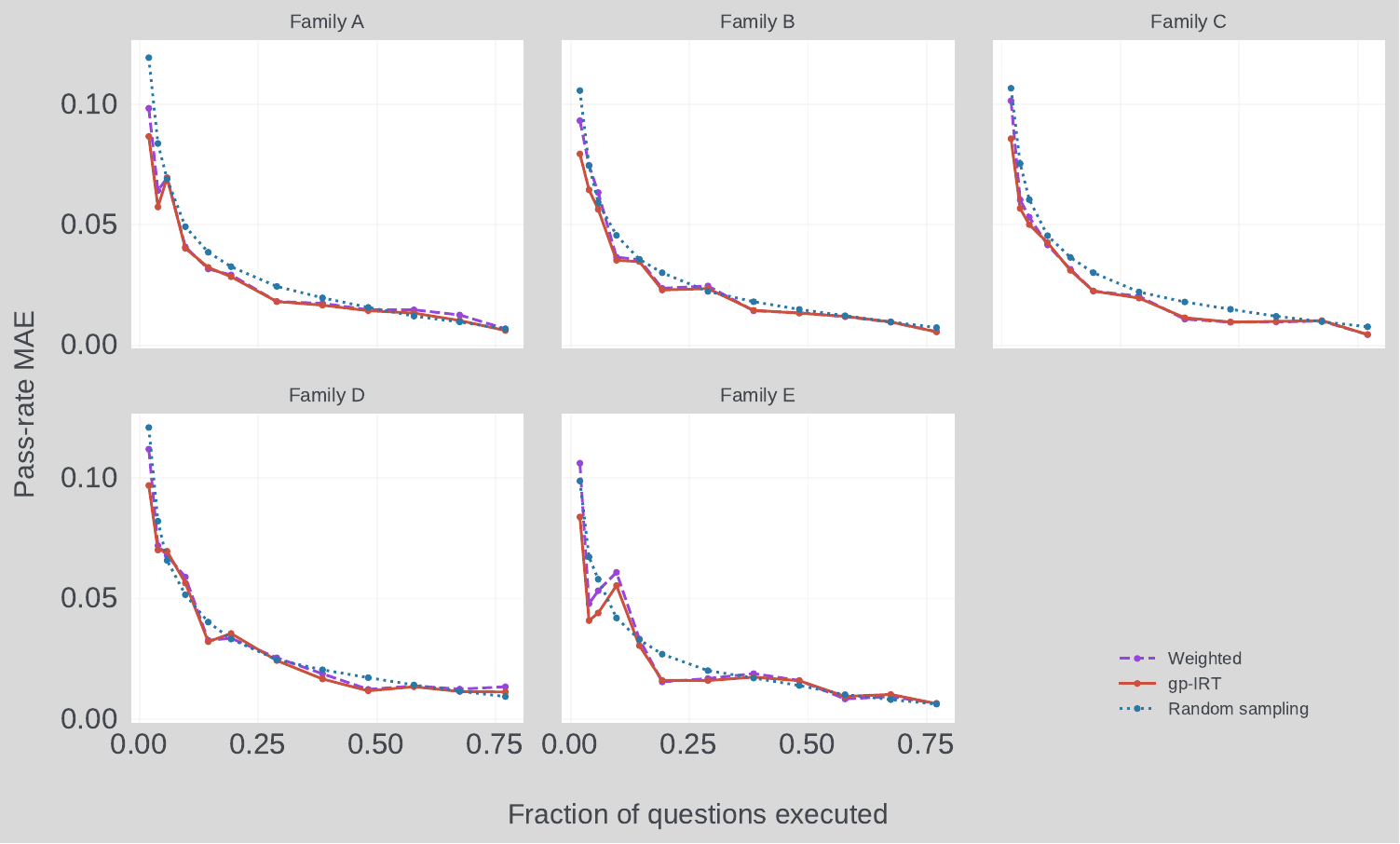}
    \caption{Cross-agent pass-rate MAE by anonymized family for the transferred
    difficulty-stratified gp-IRT subset and 20-seed random sampling. Lower is
    better. Each panel compares methods on the same held-out runs within that
    family.}
    \label{fig:cross-agent-family-mae}
\end{figure*}

\begin{figure*}[t]
    \centering
    \includegraphics[width=0.82\textwidth]{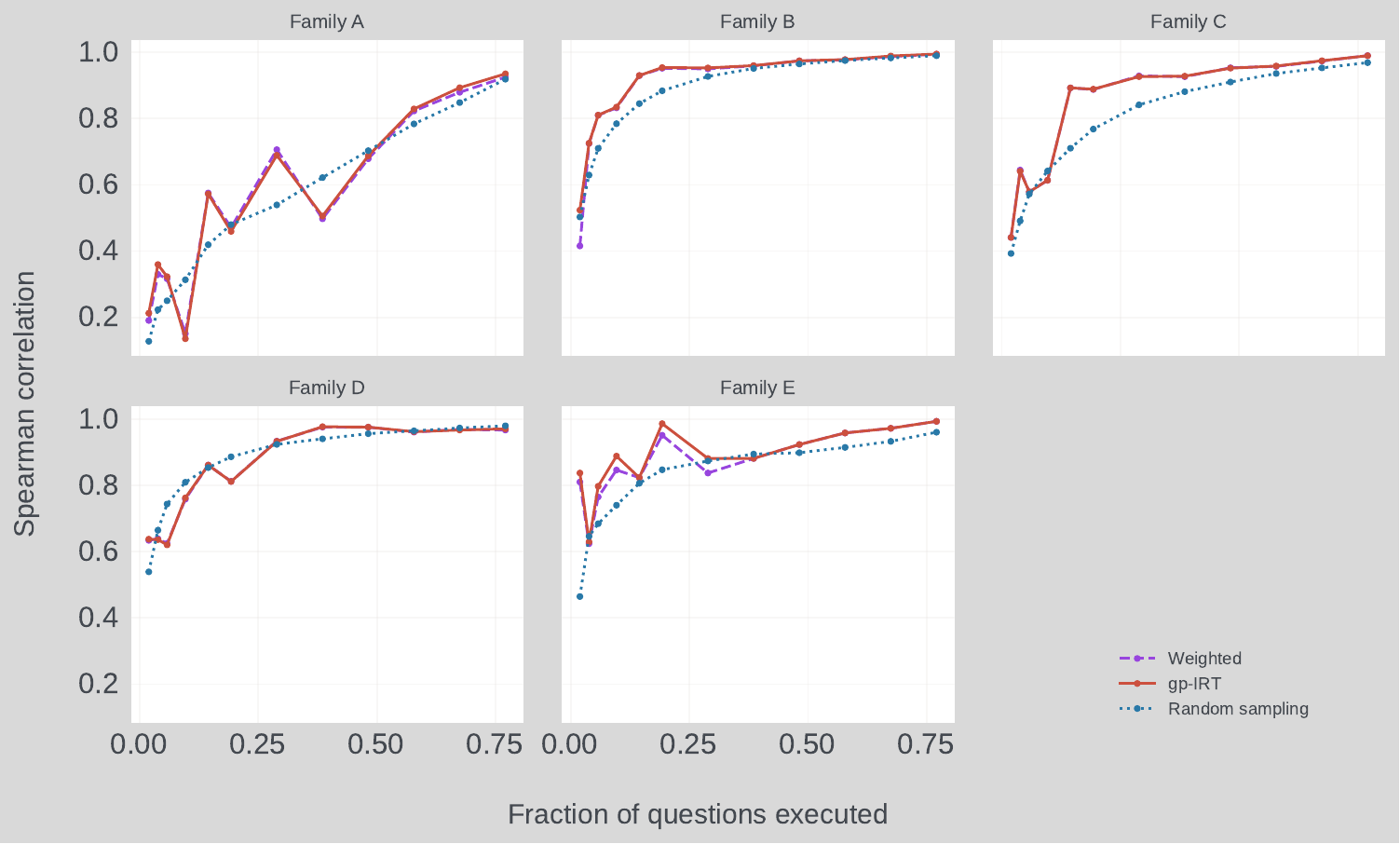}
    \caption{Cross-agent Spearman ranking fidelity by anonymized family. Higher
    is better. Correlations are computed within family, and random sampling is
    averaged over 20 seeds.}
    \label{fig:cross-agent-family-spearman}
\end{figure*}

\begin{figure*}[t]
    \centering
    \includegraphics[width=0.82\textwidth]{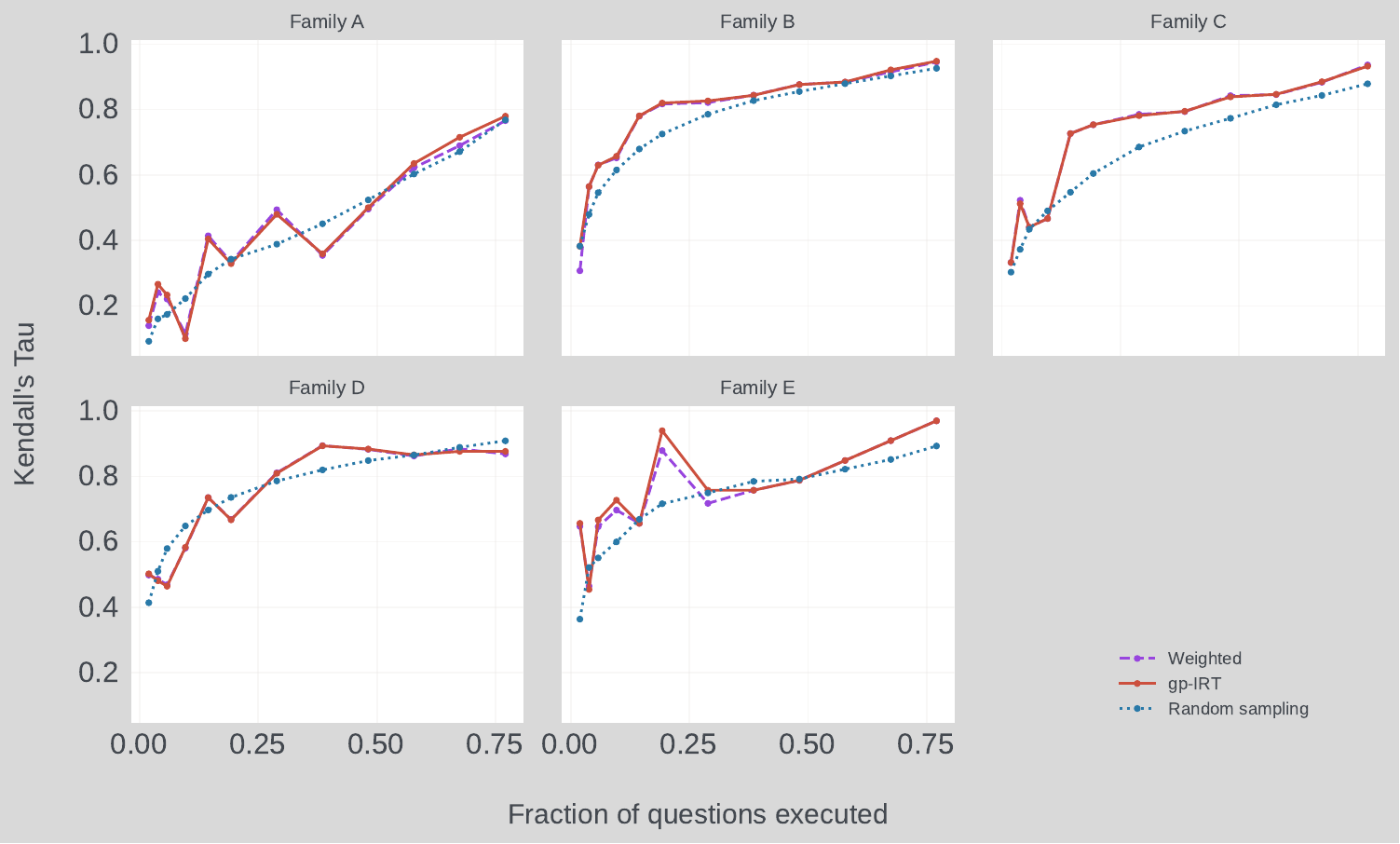}
    \caption{Cross-agent Kendall's $\tau$ ranking fidelity by anonymized
    family. Higher is better. Correlations are computed within family, and
    random sampling is averaged over 20 seeds.}
    \label{fig:cross-agent-family-kendall}
\end{figure*}

\paragraph{Cross-agent transfer results.}
Table~\ref{tab:cross-agent-run-counts} first defines the five held-out
families and their run counts. For the aggregate transfer result,
Figure~\ref{fig:cross-agent-transfer-ranking} and
Table~\ref{tab:transfer-pooled-complete} provide the pooled ranking curves
and all exact pooled metrics. 
Figures~\ref{fig:cross-agent-family-mae}--
\ref{fig:cross-agent-family-kendall} and
Tables~\ref{tab:transfer-family-a-complete}--
\ref{tab:transfer-family-e-complete} report the same results
separately for each agent family.

\begin{figure*}[t]
    \centering
    \begin{subfigure}[t]{0.49\textwidth}
        \centering
        \includegraphics[width=\linewidth]{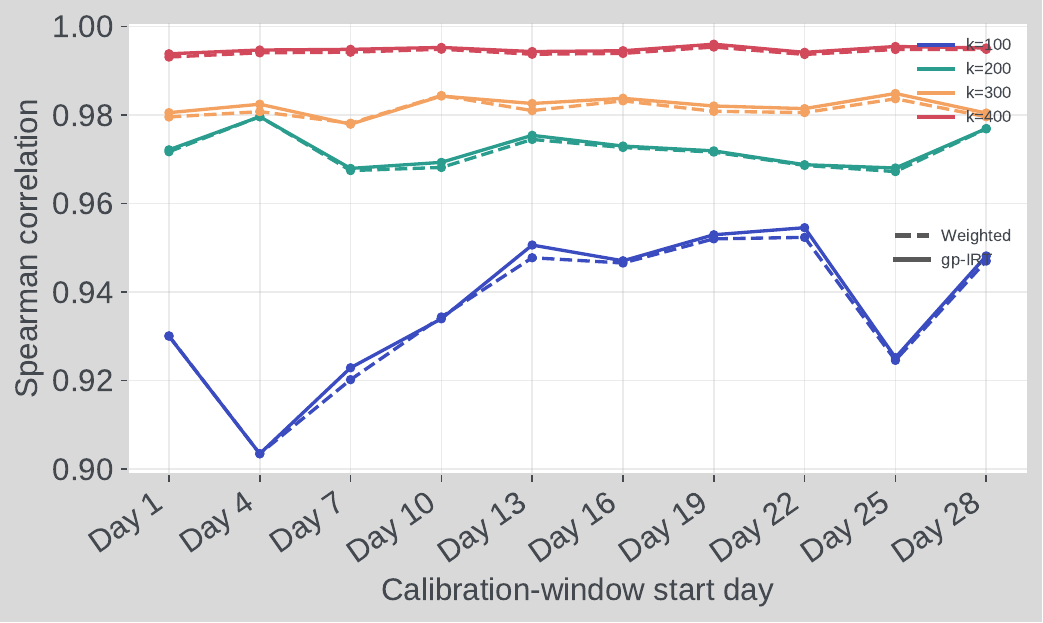}
        \caption{Spearman correlation.}
        \label{fig:calibration-window-spearman}
    \end{subfigure}
    \hfill
    \begin{subfigure}[t]{0.49\textwidth}
        \centering
        \includegraphics[width=\linewidth]{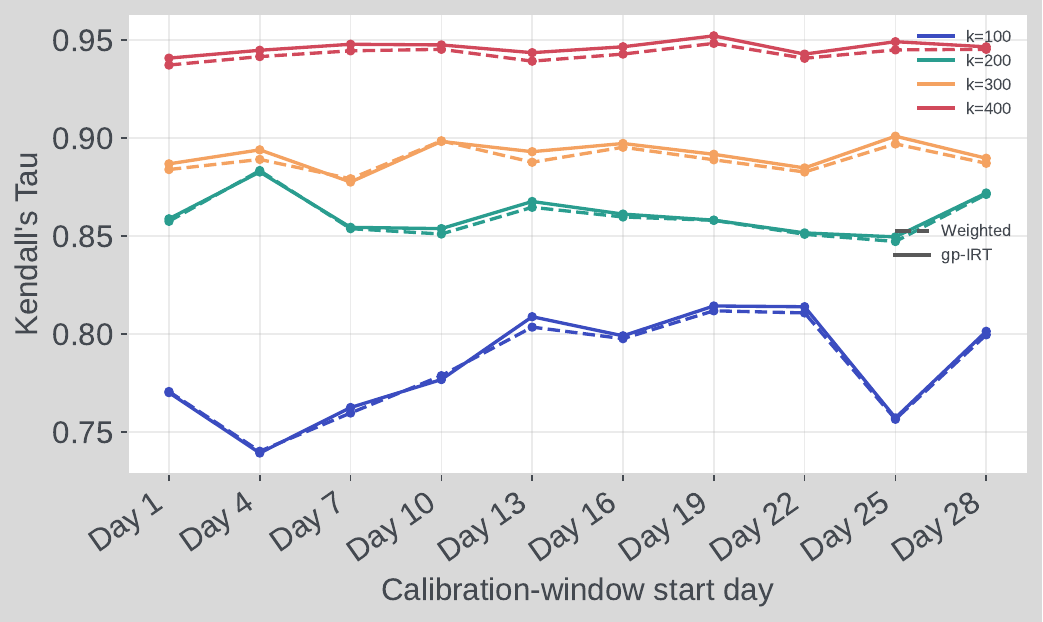}
        \caption{Kendall's $\tau$.}
        \label{fig:calibration-window-kendall}
    \end{subfigure}
    \caption{Ranking-fidelity sensitivity of difficulty-stratified fixed
    subsets to nested calibration windows ending Day 28. Higher values are
    better. Windows vary jointly in historical range and calibration-run
    count; all curves use the same 287 held-out evaluation runs.}
    \label{fig:calibration-window-ranking}
\end{figure*}

\paragraph{Calibration-window sensitivity.}
Figure~\ref{fig:calibration-window-ranking} gives the Spearman and Kendall
sensitivity curves that complement the MAE results in
Figure~\ref{fig:calibration-window-mae}. Table~\ref{tab:calibration-window-complete}
reports every window's start day, number of calibration runs, subset size, and
three fidelity metrics. 

\clearpage

\makeatletter
\setlength{\@fptop}{0pt}
\setlength{\@dblfptop}{0pt}
\makeatother

\input{tables/appendix_main_fixed_budget}

\FloatBarrier

\input{tables/appendix_cache_thresholds}

\FloatBarrier

\FloatBarrier

\input{tables/appendix_fixed_subset_design}

\FloatBarrier

\FloatBarrier

\input{tables/appendix_adaptive_model}

\FloatBarrier

\input{tables/appendix_transfer_run_counts}

\FloatBarrier

\FloatBarrier

\input{tables/appendix_transfer_pooled}

\FloatBarrier

\FloatBarrier

\input{tables/appendix_transfer_family_a}
\input{tables/appendix_transfer_family_b}
\input{tables/appendix_transfer_family_c}
\input{tables/appendix_transfer_family_d}
\input{tables/appendix_transfer_family_e}

\FloatBarrier

\FloatBarrier

\input{tables/appendix_calibration_windows}

\FloatBarrier

%% file: tables/appendix_main_fixed_budget.tex
\begin{table*}[t]
  \centering
  \small
  \setlength{\tabcolsep}{3.5pt}
  \begin{tabular}{rrrrrr}
    \toprule
    \multicolumn{6}{l}{\textit{(a) Pass-rate MAE (percentage points; lower is better)}} \\
    $k$ & Exec. (\%) & Random & \shortstack{Difficulty\\weighted} & \shortstack{Difficulty\\gp-IRT} & \shortstack{Rasch\\adaptive} \\
    \midrule
    10 & 1.9 & 10.48 & \underline{8.10} & \textbf{7.57} & 10.94 \\
    20 & 3.9 & 7.22 & \underline{6.37} & \textbf{5.51} & 9.79 \\
    30 & 5.8 & 5.86 & \underline{4.63} & \textbf{4.35} & 6.50 \\
    50 & 9.6 & 4.41 & \underline{3.37} & \textbf{3.28} & 6.46 \\
    75 & 14.5 & 3.49 & \textbf{3.04} & \underline{3.05} & 3.51 \\
    100 & 19.3 & 2.94 & \underline{2.71} & \textbf{2.65} & 2.92 \\
    150 & 28.9 & 2.25 & \underline{2.01} & \textbf{2.00} & 2.07 \\
    200 & 38.5 & 1.79 & 1.43 & \underline{1.40} & \textbf{1.37} \\
    250 & 48.2 & 1.46 & 1.17 & \underline{1.15} & \textbf{1.03} \\
    300 & 57.8 & 1.18 & 1.06 & \underline{0.99} & \textbf{0.79} \\
    350 & 67.4 & 0.93 & 0.77 & \underline{0.73} & \textbf{0.60} \\
    400 & 77.1 & 0.69 & 0.51 & \underline{0.47} & \textbf{0.42} \\
    \midrule
    \multicolumn{6}{l}{\textit{(b) Spearman correlation (higher is better)}} \\
    $k$ & Exec. (\%) & Random & \shortstack{Difficulty\\weighted} & \shortstack{Difficulty\\gp-IRT} & \shortstack{Rasch\\adaptive} \\
    \midrule
    10 & 1.9 & 0.559 & \underline{0.586} & \textbf{0.598} & 0.446 \\
    20 & 3.9 & 0.692 & \textbf{0.804} & \underline{0.803} & 0.510 \\
    30 & 5.8 & 0.757 & \textbf{0.803} & \underline{0.796} & 0.667 \\
    50 & 9.6 & 0.835 & \underline{0.876} & \textbf{0.883} & 0.697 \\
    75 & 14.5 & \underline{0.884} & \textbf{0.924} & \textbf{0.924} & 0.883 \\
    100 & 19.3 & 0.911 & \textbf{0.930} & \textbf{0.930} & \underline{0.923} \\
    150 & 28.9 & 0.941 & \underline{0.960} & \textbf{0.961} & 0.948 \\
    200 & 38.5 & \underline{0.957} & \textbf{0.972} & \textbf{0.972} & \textbf{0.972} \\
    250 & 48.2 & 0.967 & 0.977 & \underline{0.978} & \textbf{0.979} \\
    300 & 57.8 & 0.975 & 0.980 & \underline{0.981} & \textbf{0.986} \\
    350 & 67.4 & 0.983 & 0.987 & \underline{0.988} & \textbf{0.992} \\
    400 & 77.1 & 0.989 & 0.993 & \underline{0.994} & \textbf{0.995} \\
    \midrule
    \multicolumn{6}{l}{\textit{(c) Kendall's Tau (higher is better)}} \\
    $k$ & Exec. (\%) & Random & \shortstack{Difficulty\\weighted} & \shortstack{Difficulty\\gp-IRT} & \shortstack{Rasch\\adaptive} \\
    \midrule
    10 & 1.9 & 0.426 & \textbf{0.434} & \underline{0.431} & 0.323 \\
    20 & 3.9 & 0.531 & \textbf{0.632} & \underline{0.619} & 0.383 \\
    30 & 5.8 & 0.585 & \textbf{0.617} & \underline{0.607} & 0.517 \\
    50 & 9.6 & 0.661 & \underline{0.692} & \textbf{0.700} & 0.570 \\
    75 & 14.5 & 0.715 & \textbf{0.764} & \textbf{0.764} & \underline{0.719} \\
    100 & 19.3 & 0.751 & \textbf{0.771} & \underline{0.770} & 0.760 \\
    150 & 28.9 & 0.798 & \underline{0.832} & \textbf{0.834} & 0.813 \\
    200 & 38.5 & 0.829 & 0.857 & \underline{0.859} & \textbf{0.873} \\
    250 & 48.2 & 0.853 & \underline{0.874} & \underline{0.874} & \textbf{0.900} \\
    300 & 57.8 & 0.875 & 0.884 & \underline{0.887} & \textbf{0.920} \\
    350 & 67.4 & 0.898 & 0.912 & \underline{0.915} & \textbf{0.944} \\
    400 & 77.1 & 0.923 & 0.937 & \underline{0.941} & \textbf{0.962} \\
    \bottomrule
  \end{tabular}
  \caption{Complete fixed-budget headline results on 287 held-out runs. Within each budget and metric, bold marks the best displayed value and underlining marks the second-best; emphasis is descriptive and does not denote statistical significance.}
  \label{tab:main-fixed-budget-complete}
\end{table*}

%% file: tables/appendix_cache_thresholds.tex
\begin{table}[t]
  \centering
  \scriptsize
  \setlength{\tabcolsep}{2.2pt}
  \begin{tabular}{rrrrr}
    \toprule
    Threshold $q$ & Exec. (\%) & MAE (pp) & Spearman & Kendall \\
    \midrule
    0.30 & 20.3 & 7.36 & 0.904 & 0.732 \\
    0.29 & 21.4 & 7.19 & 0.913 & 0.747 \\
    0.28 & 22.4 & 7.06 & 0.918 & 0.755 \\
    0.27 & 23.7 & 6.86 & 0.922 & 0.760 \\
    0.26 & 25.0 & 6.67 & 0.925 & 0.766 \\
    0.25 & 26.2 & 6.47 & 0.932 & 0.780 \\
    0.24 & 27.5 & 6.22 & 0.933 & 0.781 \\
    0.23 & 29.1 & 5.97 & 0.938 & 0.790 \\
    0.22 & 30.5 & 5.78 & 0.942 & 0.798 \\
    0.21 & 32.2 & 5.54 & 0.948 & 0.809 \\
    0.20 & 33.6 & 5.32 & 0.953 & 0.820 \\
    0.19 & 35.5 & 5.05 & 0.957 & 0.831 \\
    0.18 & 37.0 & 4.84 & 0.961 & 0.840 \\
    0.17 & 38.7 & 4.62 & 0.963 & 0.846 \\
    0.16 & 40.1 & 4.44 & 0.967 & 0.855 \\
    0.15 & 41.9 & 4.20 & 0.969 & 0.862 \\
    0.14 & 43.7 & 4.01 & 0.972 & 0.869 \\
    0.13 & 45.6 & 3.77 & 0.974 & 0.876 \\
    0.12 & 47.2 & 3.61 & 0.977 & 0.884 \\
    0.11 & 48.9 & 3.42 & 0.979 & 0.890 \\
    0.10 & 50.7 & 3.22 & 0.981 & 0.899 \\
    0.09 & 52.9 & 2.97 & 0.982 & 0.905 \\
    0.08 & 55.1 & 2.74 & 0.983 & 0.911 \\
    0.07 & 57.6 & 2.48 & 0.984 & 0.915 \\
    0.06 & 59.7 & 2.27 & 0.985 & 0.920 \\
    0.05 & 62.4 & 2.04 & 0.986 & 0.923 \\
    0.04 & 65.4 & 1.80 & 0.989 & 0.935 \\
    0.03 & 69.1 & 1.54 & 0.992 & 0.943 \\
    0.02 & 73.5 & 1.23 & 0.995 & 0.957 \\
    0.01 & 79.0 & 0.91 & 0.996 & 0.967 \\
    0.00 & 81.6 & 0.81 & 0.997 & 0.974 \\
    \bottomrule
  \end{tabular}
  \caption{Complete historical-caching threshold sweep on 287 held-out runs. Rows are ordered by mean execution fraction. MAE is in percentage points.}
  \label{tab:cache-complete}
\end{table}

%% file: tables/appendix_fixed_subset_design.tex
\begin{table*}[t]
  \centering
  \scriptsize
  \setlength{\tabcolsep}{2.4pt}
  \begin{tabular}{rrrrrrrrr}
    \toprule
    \multicolumn{9}{l}{\textit{(a) Pass-rate MAE (percentage points; lower is better)}} \\
    $k$ & Exec. (\%) & Random & \shortstack{Difficulty\\weighted} & \shortstack{Difficulty\\gp-IRT} & \shortstack{Hist. K-means\\weighted} & \shortstack{Hist. K-means\\gp-IRT} & \shortstack{IRT K-means\\weighted} & \shortstack{IRT K-means\\gp-IRT} \\
    \midrule
    10 & 1.9 & 10.48 & 8.10 & 7.57 & \textbf{6.24} & 10.11 & 10.44 & \underline{6.53} \\
    20 & 3.9 & 7.22 & 6.37 & 5.51 & \textbf{4.51} & 5.82 & 8.00 & \underline{5.00} \\
    30 & 5.8 & 5.86 & 4.63 & \underline{4.35} & 4.37 & \textbf{3.88} & 9.10 & 5.27 \\
    50 & 9.6 & 4.41 & 3.37 & \underline{3.28} & 3.44 & \textbf{2.77} & 7.95 & 5.47 \\
    75 & 14.5 & 3.49 & \underline{3.04} & 3.05 & 4.12 & \textbf{2.96} & 4.76 & 3.29 \\
    100 & 19.3 & 2.94 & 2.71 & \underline{2.65} & 2.93 & \textbf{2.46} & 4.21 & 2.88 \\
    150 & 28.9 & 2.25 & \underline{2.01} & \textbf{2.00} & 2.61 & 2.13 & 3.36 & 2.71 \\
    200 & 38.5 & 1.79 & \underline{1.43} & \textbf{1.40} & 2.51 & 2.12 & 2.47 & 2.07 \\
    250 & 48.2 & 1.46 & \underline{1.17} & \textbf{1.15} & 1.75 & 1.56 & 2.06 & 1.70 \\
    300 & 57.8 & 1.18 & \underline{1.06} & \textbf{0.99} & 1.38 & 1.27 & 1.63 & 1.50 \\
    350 & 67.4 & 0.93 & \underline{0.77} & \textbf{0.73} & 1.18 & 1.05 & 1.52 & 1.42 \\
    400 & 77.1 & 0.69 & \underline{0.51} & \textbf{0.47} & 1.07 & 0.93 & 0.94 & 0.88 \\
    \midrule
    \multicolumn{9}{l}{\textit{(b) Spearman correlation (higher is better)}} \\
    $k$ & Exec. (\%) & Random & \shortstack{Difficulty\\weighted} & \shortstack{Difficulty\\gp-IRT} & \shortstack{Hist. K-means\\weighted} & \shortstack{Hist. K-means\\gp-IRT} & \shortstack{IRT K-means\\weighted} & \shortstack{IRT K-means\\gp-IRT} \\
    \midrule
    10 & 1.9 & 0.559 & 0.586 & 0.598 & \textbf{0.823} & \underline{0.765} & 0.394 & 0.555 \\
    20 & 3.9 & 0.692 & \underline{0.804} & 0.803 & \textbf{0.862} & 0.796 & 0.641 & 0.752 \\
    30 & 5.8 & 0.757 & 0.803 & 0.796 & \underline{0.852} & \textbf{0.863} & 0.559 & 0.722 \\
    50 & 9.6 & 0.835 & 0.876 & 0.883 & \underline{0.906} & \textbf{0.930} & 0.769 & 0.861 \\
    75 & 14.5 & 0.884 & \underline{0.924} & \underline{0.924} & 0.921 & \textbf{0.929} & 0.873 & 0.903 \\
    100 & 19.3 & 0.911 & \underline{0.930} & \underline{0.930} & 0.924 & \textbf{0.934} & 0.872 & 0.907 \\
    150 & 28.9 & 0.941 & \underline{0.960} & \textbf{0.961} & 0.944 & 0.953 & 0.908 & 0.932 \\
    200 & 38.5 & 0.957 & \textbf{0.972} & \textbf{0.972} & 0.948 & \underline{0.958} & 0.925 & 0.940 \\
    250 & 48.2 & 0.967 & \underline{0.977} & \textbf{0.978} & 0.969 & 0.974 & 0.949 & 0.957 \\
    300 & 57.8 & 0.975 & 0.980 & \underline{0.981} & 0.979 & \textbf{0.982} & 0.966 & 0.971 \\
    350 & 67.4 & 0.983 & 0.987 & \underline{0.988} & 0.987 & \textbf{0.989} & 0.968 & 0.973 \\
    400 & 77.1 & 0.989 & \underline{0.993} & \textbf{0.994} & \underline{0.993} & \textbf{0.994} & 0.984 & 0.986 \\
    \midrule
    \multicolumn{9}{l}{\textit{(c) Kendall's Tau (higher is better)}} \\
    $k$ & Exec. (\%) & Random & \shortstack{Difficulty\\weighted} & \shortstack{Difficulty\\gp-IRT} & \shortstack{Hist. K-means\\weighted} & \shortstack{Hist. K-means\\gp-IRT} & \shortstack{IRT K-means\\weighted} & \shortstack{IRT K-means\\gp-IRT} \\
    \midrule
    10 & 1.9 & 0.426 & 0.434 & 0.431 & \textbf{0.611} & \underline{0.566} & 0.282 & 0.399 \\
    20 & 3.9 & 0.531 & \underline{0.632} & 0.619 & \textbf{0.668} & 0.598 & 0.457 & 0.557 \\
    30 & 5.8 & 0.585 & 0.617 & 0.607 & \underline{0.656} & \textbf{0.668} & 0.397 & 0.528 \\
    50 & 9.6 & 0.661 & 0.692 & 0.700 & \underline{0.728} & \textbf{0.766} & 0.580 & 0.676 \\
    75 & 14.5 & 0.715 & \textbf{0.764} & \textbf{0.764} & \underline{0.749} & \textbf{0.764} & 0.682 & 0.724 \\
    100 & 19.3 & 0.751 & \underline{0.771} & 0.770 & 0.762 & \textbf{0.775} & 0.693 & 0.737 \\
    150 & 28.9 & 0.798 & \underline{0.832} & \textbf{0.834} & 0.804 & 0.820 & 0.733 & 0.772 \\
    200 & 38.5 & 0.829 & \underline{0.857} & \textbf{0.859} & 0.811 & 0.830 & 0.760 & 0.790 \\
    250 & 48.2 & 0.853 & \textbf{0.874} & \textbf{0.874} & \underline{0.867} & \textbf{0.874} & 0.806 & 0.825 \\
    300 & 57.8 & 0.875 & 0.884 & 0.887 & \underline{0.893} & \textbf{0.897} & 0.846 & 0.859 \\
    350 & 67.4 & 0.898 & 0.912 & \underline{0.915} & 0.913 & \textbf{0.920} & 0.851 & 0.865 \\
    400 & 77.1 & 0.923 & 0.937 & \underline{0.941} & 0.940 & \textbf{0.944} & 0.896 & 0.902 \\
    \bottomrule
  \end{tabular}
  \caption{Complete fixed-subset selection and estimation results on 287 held-out runs. Bold and underlining mark the best and second-best displayed values at each budget, respectively.}
  \label{tab:fixed-design-complete}
\end{table*}

%% file: tables/appendix_adaptive_model.tex
\begin{table*}[t]
  \centering
  \small
  \setlength{\tabcolsep}{3.5pt}
  \begin{tabular}{rrrrr}
    \toprule
    \multicolumn{5}{l}{\textit{(a) Pass-rate MAE (percentage points; lower is better)}} \\
    $k$ & Exec. (\%) & Random & \shortstack{Rasch\\adaptive} & \shortstack{Multidimensional 2PL\\adaptive} \\
    \midrule
    10 & 1.9 & \underline{10.48} & 10.94 & \textbf{5.07} \\
    20 & 3.9 & \underline{7.22} & 9.79 & \textbf{3.90} \\
    30 & 5.8 & \underline{5.86} & 6.50 & \textbf{3.27} \\
    50 & 9.6 & \underline{4.41} & 6.46 & \textbf{2.81} \\
    75 & 14.5 & \underline{3.49} & 3.51 & \textbf{2.39} \\
    100 & 19.3 & 2.94 & \underline{2.92} & \textbf{1.97} \\
    150 & 28.9 & 2.25 & \underline{2.07} & \textbf{1.44} \\
    200 & 38.5 & 1.79 & \underline{1.37} & \textbf{1.03} \\
    250 & 48.2 & 1.46 & \underline{1.03} & \textbf{0.73} \\
    300 & 57.8 & 1.18 & \underline{0.79} & \textbf{0.58} \\
    350 & 67.4 & 0.93 & \underline{0.60} & \textbf{0.46} \\
    400 & 77.1 & 0.69 & \underline{0.42} & \textbf{0.29} \\
    \midrule
    \multicolumn{5}{l}{\textit{(b) Spearman correlation (higher is better)}} \\
    $k$ & Exec. (\%) & Random & \shortstack{Rasch\\adaptive} & \shortstack{Multidimensional 2PL\\adaptive} \\
    \midrule
    10 & 1.9 & \underline{0.559} & 0.446 & \textbf{0.762} \\
    20 & 3.9 & \underline{0.692} & 0.510 & \textbf{0.832} \\
    30 & 5.8 & \underline{0.757} & 0.667 & \textbf{0.884} \\
    50 & 9.6 & \underline{0.835} & 0.697 & \textbf{0.908} \\
    75 & 14.5 & \underline{0.884} & 0.883 & \textbf{0.929} \\
    100 & 19.3 & 0.911 & \underline{0.923} & \textbf{0.954} \\
    150 & 28.9 & 0.941 & \underline{0.948} & \textbf{0.972} \\
    200 & 38.5 & 0.957 & \underline{0.972} & \textbf{0.982} \\
    250 & 48.2 & 0.967 & \underline{0.979} & \textbf{0.989} \\
    300 & 57.8 & 0.975 & \underline{0.986} & \textbf{0.992} \\
    350 & 67.4 & 0.983 & \underline{0.992} & \textbf{0.995} \\
    400 & 77.1 & 0.989 & \underline{0.995} & \textbf{0.998} \\
    \midrule
    \multicolumn{5}{l}{\textit{(c) Kendall's Tau (higher is better)}} \\
    $k$ & Exec. (\%) & Random & \shortstack{Rasch\\adaptive} & \shortstack{Multidimensional 2PL\\adaptive} \\
    \midrule
    10 & 1.9 & \underline{0.426} & 0.323 & \textbf{0.575} \\
    20 & 3.9 & \underline{0.531} & 0.383 & \textbf{0.640} \\
    30 & 5.8 & \underline{0.585} & 0.517 & \textbf{0.702} \\
    50 & 9.6 & \underline{0.661} & 0.570 & \textbf{0.738} \\
    75 & 14.5 & 0.715 & \underline{0.719} & \textbf{0.774} \\
    100 & 19.3 & 0.751 & \underline{0.760} & \textbf{0.819} \\
    150 & 28.9 & 0.798 & \underline{0.813} & \textbf{0.864} \\
    200 & 38.5 & 0.829 & \underline{0.873} & \textbf{0.894} \\
    250 & 48.2 & 0.853 & \underline{0.900} & \textbf{0.918} \\
    300 & 57.8 & 0.875 & \underline{0.920} & \textbf{0.933} \\
    350 & 67.4 & 0.898 & \underline{0.944} & \textbf{0.948} \\
    400 & 77.1 & 0.923 & \underline{0.962} & \textbf{0.970} \\
    \bottomrule
  \end{tabular}
  \caption{Complete adaptive-model comparison on 287 held-out runs. Bold and underlining mark the best and second-best displayed values at each budget, respectively.}
  \label{tab:adaptive-model-complete}
\end{table*}

%% file: tables/appendix_transfer_run_counts.tex
\begin{table}[t]
  \centering
  \small
  \begin{tabular}{lr}
    \toprule
    Agent family & Evaluation runs \\
    \midrule
    Family A & 51 \\
    Family B & 146 \\
    Family C & 56 \\
    Family D & 34 \\
    Family E & 12 \\
    \midrule
    Total & 299 \\
    \bottomrule
  \end{tabular}
  \caption{Breakdown of transferred evaluation runs by anonymized agent family. The labels match the family-level figures and tables.}
  \label{tab:cross-agent-run-counts}
\end{table}

%% file: tables/appendix_transfer_pooled.tex
\begin{table*}[t]
  \centering
  \scriptsize
  \setlength{\tabcolsep}{3.5pt}
  \begin{tabular}{rrrrrrr}
    \toprule
    \multicolumn{7}{l}{\textit{(a) Pass-rate MAE (percentage points; lower is better)}} \\
    $k$ & Exec. (\%) & \shortstack{Original\\weighted} & \shortstack{Original\\gp-IRT} & \shortstack{Transfer\\random} & \shortstack{Transfer\\weighted} & \shortstack{Transfer\\gp-IRT} \\
    \midrule
    10 & 1.9 & \underline{8.10} & \textbf{7.57} & 10.96 & \underline{9.83} & \textbf{8.40} \\
    20 & 3.9 & \underline{6.37} & \textbf{5.51} & 7.69 & \underline{6.88} & \textbf{6.15} \\
    30 & 5.8 & \underline{4.63} & \textbf{4.35} & \underline{6.19} & 6.26 & \textbf{5.85} \\
    50 & 9.6 & \underline{3.37} & \textbf{3.28} & 4.67 & \underline{4.18} & \textbf{4.07} \\
    75 & 14.5 & \textbf{3.04} & \underline{3.05} & 3.68 & \underline{3.37} & \textbf{3.32} \\
    100 & 19.3 & \underline{2.71} & \textbf{2.65} & 3.08 & \underline{2.52} & \textbf{2.50} \\
    150 & 28.9 & \underline{2.01} & \textbf{2.00} & 2.29 & \underline{2.25} & \textbf{2.17} \\
    200 & 38.5 & \underline{1.43} & \textbf{1.40} & 1.86 & \underline{1.49} & \textbf{1.47} \\
    250 & 48.2 & \underline{1.17} & \textbf{1.15} & 1.53 & \underline{1.30} & \textbf{1.28} \\
    300 & 57.8 & \underline{1.06} & \textbf{0.99} & 1.24 & \underline{1.21} & \textbf{1.19} \\
    350 & 67.4 & \underline{0.77} & \textbf{0.73} & \textbf{0.99} & 1.05 & \underline{1.02} \\
    400 & 77.1 & \underline{0.51} & \textbf{0.47} & 0.75 & \underline{0.66} & \textbf{0.62} \\
    \midrule
    \multicolumn{7}{l}{\textit{(b) Spearman correlation (higher is better)}} \\
    $k$ & Exec. (\%) & \shortstack{Original\\weighted} & \shortstack{Original\\gp-IRT} & \shortstack{Transfer\\random} & \shortstack{Transfer\\weighted} & \shortstack{Transfer\\gp-IRT} \\
    \midrule
    10 & 1.9 & \underline{0.586} & \textbf{0.598} & 0.600 & \underline{0.648} & \textbf{0.683} \\
    20 & 3.9 & \textbf{0.804} & \underline{0.803} & \underline{0.702} & \textbf{0.817} & \textbf{0.817} \\
    30 & 5.8 & \textbf{0.803} & \underline{0.796} & 0.756 & \underline{0.844} & \textbf{0.847} \\
    50 & 9.6 & \underline{0.876} & \textbf{0.883} & 0.821 & \textbf{0.853} & \underline{0.846} \\
    75 & 14.5 & \textbf{0.924} & \textbf{0.924} & 0.869 & \textbf{0.937} & \underline{0.936} \\
    100 & 19.3 & \textbf{0.930} & \textbf{0.930} & 0.900 & \underline{0.947} & \textbf{0.948} \\
    150 & 28.9 & \underline{0.960} & \textbf{0.961} & 0.935 & \underline{0.938} & \textbf{0.941} \\
    200 & 38.5 & \textbf{0.972} & \textbf{0.972} & \underline{0.955} & \textbf{0.965} & \textbf{0.965} \\
    250 & 48.2 & \underline{0.977} & \textbf{0.978} & \underline{0.968} & \textbf{0.974} & \textbf{0.974} \\
    300 & 57.8 & \underline{0.980} & \textbf{0.981} & \underline{0.979} & \textbf{0.981} & \textbf{0.981} \\
    350 & 67.4 & \underline{0.987} & \textbf{0.988} & 0.986 & \underline{0.987} & \textbf{0.988} \\
    400 & 77.1 & \underline{0.993} & \textbf{0.994} & \underline{0.992} & \textbf{0.995} & \textbf{0.995} \\
    \midrule
    \multicolumn{7}{l}{\textit{(c) Kendall's Tau (higher is better)}} \\
    $k$ & Exec. (\%) & \shortstack{Original\\weighted} & \shortstack{Original\\gp-IRT} & \shortstack{Transfer\\random} & \shortstack{Transfer\\weighted} & \shortstack{Transfer\\gp-IRT} \\
    \midrule
    10 & 1.9 & \textbf{0.434} & \underline{0.431} & 0.465 & \underline{0.497} & \textbf{0.522} \\
    20 & 3.9 & \textbf{0.632} & \underline{0.619} & 0.549 & \textbf{0.655} & \underline{0.652} \\
    30 & 5.8 & \textbf{0.617} & \underline{0.607} & 0.597 & \underline{0.672} & \textbf{0.673} \\
    50 & 9.6 & \underline{0.692} & \textbf{0.700} & 0.661 & \textbf{0.680} & \underline{0.675} \\
    75 & 14.5 & \textbf{0.764} & \textbf{0.764} & 0.713 & \textbf{0.799} & \underline{0.797} \\
    100 & 19.3 & \textbf{0.771} & \underline{0.770} & 0.752 & \underline{0.816} & \textbf{0.818} \\
    150 & 28.9 & \underline{0.832} & \textbf{0.834} & 0.804 & \underline{0.808} & \textbf{0.812} \\
    200 & 38.5 & \underline{0.857} & \textbf{0.859} & 0.838 & \textbf{0.858} & \underline{0.856} \\
    250 & 48.2 & \textbf{0.874} & \textbf{0.874} & 0.865 & \textbf{0.879} & \underline{0.878} \\
    300 & 57.8 & \underline{0.884} & \textbf{0.887} & \underline{0.891} & \underline{0.891} & \textbf{0.892} \\
    350 & 67.4 & \underline{0.912} & \textbf{0.915} & 0.911 & \underline{0.912} & \textbf{0.917} \\
    400 & 77.1 & \underline{0.937} & \textbf{0.941} & 0.934 & \underline{0.946} & \textbf{0.949} \\
    \bottomrule
  \end{tabular}
  \caption{Complete pooled cross-agent transfer results. Original denotes the 287 calibration-agent test runs; Transfer denotes 299 pooled runs from five additional families. Emphasis is computed separately within the original and transferred populations.}
  \label{tab:transfer-pooled-complete}
\end{table*}

%% file: tables/appendix_transfer_family_a.tex
\begin{table*}[t]
  \centering
  \small
  \setlength{\tabcolsep}{3.5pt}
  \begin{tabular}{rrrrr}
    \toprule
    \multicolumn{5}{l}{\textit{(a) Pass-rate MAE (percentage points; lower is better)}} \\
    $k$ & Exec. (\%) & Random & Weighted & gp-IRT \\
    \midrule
    10 & 1.9 & 11.93 & \underline{9.83} & \textbf{8.66} \\
    20 & 3.9 & 8.37 & \underline{6.42} & \textbf{5.74} \\
    30 & 5.8 & \textbf{6.90} & \underline{6.95} & \underline{6.95} \\
    50 & 9.6 & 4.92 & \underline{4.08} & \textbf{4.02} \\
    75 & 14.5 & 3.86 & \textbf{3.17} & \underline{3.23} \\
    100 & 19.3 & 3.26 & \underline{2.92} & \textbf{2.84} \\
    150 & 28.9 & \underline{2.45} & \textbf{1.82} & \textbf{1.82} \\
    200 & 38.5 & 1.97 & \underline{1.74} & \textbf{1.67} \\
    250 & 48.2 & 1.58 & \underline{1.47} & \textbf{1.44} \\
    300 & 57.8 & \textbf{1.21} & 1.47 & \underline{1.34} \\
    350 & 67.4 & \textbf{0.97} & 1.26 & \underline{1.03} \\
    400 & 77.1 & \underline{0.68} & 0.69 & \textbf{0.61} \\
    \midrule
    \multicolumn{5}{l}{\textit{(b) Spearman correlation (higher is better)}} \\
    $k$ & Exec. (\%) & Random & Weighted & gp-IRT \\
    \midrule
    10 & 1.9 & 0.128 & \underline{0.191} & \textbf{0.213} \\
    20 & 3.9 & 0.223 & \underline{0.330} & \textbf{0.360} \\
    30 & 5.8 & 0.250 & \underline{0.317} & \textbf{0.323} \\
    50 & 9.6 & \textbf{0.314} & \underline{0.151} & 0.136 \\
    75 & 14.5 & 0.419 & \textbf{0.575} & \underline{0.573} \\
    100 & 19.3 & \textbf{0.480} & \underline{0.471} & 0.460 \\
    150 & 28.9 & 0.539 & \textbf{0.706} & \underline{0.689} \\
    200 & 38.5 & \textbf{0.622} & 0.498 & \underline{0.506} \\
    250 & 48.2 & \textbf{0.703} & 0.679 & \underline{0.687} \\
    300 & 57.8 & 0.784 & \underline{0.823} & \textbf{0.829} \\
    350 & 67.4 & 0.848 & \underline{0.879} & \textbf{0.893} \\
    400 & 77.1 & 0.919 & \underline{0.925} & \textbf{0.935} \\
    \midrule
    \multicolumn{5}{l}{\textit{(c) Kendall's Tau (higher is better)}} \\
    $k$ & Exec. (\%) & Random & Weighted & gp-IRT \\
    \midrule
    10 & 1.9 & 0.092 & \underline{0.140} & \textbf{0.157} \\
    20 & 3.9 & 0.161 & \underline{0.241} & \textbf{0.267} \\
    30 & 5.8 & 0.174 & \underline{0.222} & \textbf{0.234} \\
    50 & 9.6 & \textbf{0.223} & \underline{0.115} & 0.100 \\
    75 & 14.5 & 0.298 & \textbf{0.415} & \underline{0.405} \\
    100 & 19.3 & \textbf{0.343} & \underline{0.334} & 0.330 \\
    150 & 28.9 & 0.389 & \textbf{0.494} & \underline{0.480} \\
    200 & 38.5 & \textbf{0.451} & 0.355 & \underline{0.359} \\
    250 & 48.2 & \textbf{0.524} & 0.496 & \underline{0.501} \\
    300 & 57.8 & 0.603 & \underline{0.622} & \textbf{0.636} \\
    350 & 67.4 & 0.672 & \underline{0.690} & \textbf{0.716} \\
    400 & 77.1 & \underline{0.769} & 0.766 & \textbf{0.780} \\
    \bottomrule
  \end{tabular}
  \caption{Complete transfer results for anonymized Family A. Bold and underlining mark the best and second-best displayed values at each budget, respectively.}
  \label{tab:transfer-family-a-complete}
\end{table*}

%% file: tables/appendix_transfer_family_b.tex
\begin{table*}[t]
  \centering
  \small
  \setlength{\tabcolsep}{3.5pt}
  \begin{tabular}{rrrrr}
    \toprule
    \multicolumn{5}{l}{\textit{(a) Pass-rate MAE (percentage points; lower is better)}} \\
    $k$ & Exec. (\%) & Random & Weighted & gp-IRT \\
    \midrule
    10 & 1.9 & 10.56 & \underline{9.32} & \textbf{7.94} \\
    20 & 3.9 & \underline{7.46} & \underline{7.46} & \textbf{6.45} \\
    30 & 5.8 & \underline{5.93} & 6.34 & \textbf{5.64} \\
    50 & 9.6 & 4.56 & \underline{3.66} & \textbf{3.53} \\
    75 & 14.5 & 3.57 & \underline{3.56} & \textbf{3.47} \\
    100 & 19.3 & 3.01 & \underline{2.37} & \textbf{2.30} \\
    150 & 28.9 & \textbf{2.24} & 2.47 & \underline{2.35} \\
    200 & 38.5 & 1.81 & \textbf{1.43} & \underline{1.46} \\
    250 & 48.2 & 1.49 & \underline{1.35} & \textbf{1.34} \\
    300 & 57.8 & 1.23 & \textbf{1.19} & \underline{1.20} \\
    350 & 67.4 & \underline{0.97} & \textbf{0.96} & \underline{0.97} \\
    400 & 77.1 & 0.74 & \underline{0.57} & \textbf{0.56} \\
    \midrule
    \multicolumn{5}{l}{\textit{(b) Spearman correlation (higher is better)}} \\
    $k$ & Exec. (\%) & Random & Weighted & gp-IRT \\
    \midrule
    10 & 1.9 & \underline{0.503} & 0.416 & \textbf{0.524} \\
    20 & 3.9 & 0.629 & \underline{0.725} & \textbf{0.726} \\
    30 & 5.8 & \underline{0.710} & \textbf{0.810} & \textbf{0.810} \\
    50 & 9.6 & 0.785 & \underline{0.832} & \textbf{0.835} \\
    75 & 14.5 & 0.845 & \underline{0.929} & \textbf{0.930} \\
    100 & 19.3 & 0.884 & \underline{0.952} & \textbf{0.954} \\
    150 & 28.9 & 0.927 & \underline{0.950} & \textbf{0.953} \\
    200 & 38.5 & 0.951 & \underline{0.959} & \textbf{0.960} \\
    250 & 48.2 & \underline{0.965} & \textbf{0.974} & \textbf{0.974} \\
    300 & 57.8 & 0.975 & \underline{0.977} & \textbf{0.978} \\
    350 & 67.4 & 0.983 & \underline{0.987} & \textbf{0.989} \\
    400 & 77.1 & 0.990 & \underline{0.994} & \textbf{0.995} \\
    \midrule
    \multicolumn{5}{l}{\textit{(c) Kendall's Tau (higher is better)}} \\
    $k$ & Exec. (\%) & Random & Weighted & gp-IRT \\
    \midrule
    10 & 1.9 & \textbf{0.383} & 0.308 & \underline{0.382} \\
    20 & 3.9 & \underline{0.480} & \textbf{0.565} & \textbf{0.565} \\
    30 & 5.8 & 0.546 & \textbf{0.631} & \underline{0.630} \\
    50 & 9.6 & 0.615 & \underline{0.652} & \textbf{0.657} \\
    75 & 14.5 & 0.680 & \underline{0.780} & \textbf{0.781} \\
    100 & 19.3 & 0.726 & \underline{0.817} & \textbf{0.820} \\
    150 & 28.9 & 0.786 & \underline{0.821} & \textbf{0.827} \\
    200 & 38.5 & \underline{0.827} & \textbf{0.844} & \textbf{0.844} \\
    250 & 48.2 & 0.855 & \textbf{0.877} & \underline{0.876} \\
    300 & 57.8 & \underline{0.879} & \textbf{0.884} & \textbf{0.884} \\
    350 & 67.4 & 0.903 & \underline{0.915} & \textbf{0.921} \\
    400 & 77.1 & 0.926 & \underline{0.945} & \textbf{0.948} \\
    \bottomrule
  \end{tabular}
  \caption{Complete transfer results for anonymized Family B. Bold and underlining mark the best and second-best displayed values at each budget, respectively.}
  \label{tab:transfer-family-b-complete}
\end{table*}

%% file: tables/appendix_transfer_family_c.tex
\begin{table*}[t]
  \centering
  \small
  \setlength{\tabcolsep}{3.5pt}
  \begin{tabular}{rrrrr}
    \toprule
    \multicolumn{5}{l}{\textit{(a) Pass-rate MAE (percentage points; lower is better)}} \\
    $k$ & Exec. (\%) & Random & Weighted & gp-IRT \\
    \midrule
    10 & 1.9 & 10.66 & \underline{10.14} & \textbf{8.57} \\
    20 & 3.9 & 7.54 & \underline{6.03} & \textbf{5.68} \\
    30 & 5.8 & 6.04 & \underline{5.31} & \textbf{5.01} \\
    50 & 9.6 & 4.55 & \textbf{4.16} & \underline{4.26} \\
    75 & 14.5 & 3.65 & \underline{3.16} & \textbf{3.11} \\
    100 & 19.3 & \underline{3.01} & \textbf{2.25} & \textbf{2.25} \\
    150 & 28.9 & 2.21 & \underline{2.04} & \textbf{1.96} \\
    200 & 38.5 & 1.80 & \textbf{1.09} & \underline{1.15} \\
    250 & 48.2 & 1.50 & \textbf{0.96} & \underline{0.97} \\
    300 & 57.8 & 1.21 & \textbf{0.97} & \underline{0.99} \\
    350 & 67.4 & \textbf{0.98} & \underline{1.00} & 1.03 \\
    400 & 77.1 & \underline{0.77} & \textbf{0.45} & \textbf{0.45} \\
    \midrule
    \multicolumn{5}{l}{\textit{(b) Spearman correlation (higher is better)}} \\
    $k$ & Exec. (\%) & Random & Weighted & gp-IRT \\
    \midrule
    10 & 1.9 & \underline{0.393} & \textbf{0.441} & \textbf{0.441} \\
    20 & 3.9 & 0.491 & \textbf{0.645} & \underline{0.641} \\
    30 & 5.8 & \underline{0.572} & \textbf{0.579} & \textbf{0.579} \\
    50 & 9.6 & \textbf{0.642} & \underline{0.614} & \underline{0.614} \\
    75 & 14.5 & \underline{0.711} & \textbf{0.893} & \textbf{0.893} \\
    100 & 19.3 & 0.768 & \underline{0.888} & \textbf{0.889} \\
    150 & 28.9 & 0.841 & \textbf{0.929} & \underline{0.927} \\
    200 & 38.5 & 0.881 & \underline{0.926} & \textbf{0.928} \\
    250 & 48.2 & 0.910 & \textbf{0.953} & \underline{0.952} \\
    300 & 57.8 & 0.936 & \underline{0.958} & \textbf{0.959} \\
    350 & 67.4 & 0.953 & \underline{0.973} & \textbf{0.974} \\
    400 & 77.1 & \underline{0.969} & \textbf{0.990} & \textbf{0.990} \\
    \midrule
    \multicolumn{5}{l}{\textit{(c) Kendall's Tau (higher is better)}} \\
    $k$ & Exec. (\%) & Random & Weighted & gp-IRT \\
    \midrule
    10 & 1.9 & \underline{0.303} & \textbf{0.333} & \textbf{0.333} \\
    20 & 3.9 & 0.373 & \textbf{0.523} & \underline{0.513} \\
    30 & 5.8 & \underline{0.435} & \textbf{0.441} & \textbf{0.441} \\
    50 & 9.6 & \textbf{0.491} & \underline{0.467} & \underline{0.467} \\
    75 & 14.5 & \underline{0.548} & \textbf{0.727} & \textbf{0.727} \\
    100 & 19.3 & \underline{0.605} & \textbf{0.754} & \textbf{0.754} \\
    150 & 28.9 & 0.686 & \textbf{0.786} & \underline{0.781} \\
    200 & 38.5 & 0.734 & \underline{0.794} & \textbf{0.795} \\
    250 & 48.2 & 0.773 & \textbf{0.843} & \underline{0.839} \\
    300 & 57.8 & 0.815 & \underline{0.846} & \textbf{0.847} \\
    350 & 67.4 & 0.843 & \underline{0.884} & \textbf{0.885} \\
    400 & 77.1 & 0.879 & \textbf{0.936} & \underline{0.933} \\
    \bottomrule
  \end{tabular}
  \caption{Complete transfer results for anonymized Family C. Bold and underlining mark the best and second-best displayed values at each budget, respectively.}
  \label{tab:transfer-family-c-complete}
\end{table*}

%% file: tables/appendix_transfer_family_d.tex
\begin{table*}[t]
  \centering
  \small
  \setlength{\tabcolsep}{3.5pt}
  \begin{tabular}{rrrrr}
    \toprule
    \multicolumn{5}{l}{\textit{(a) Pass-rate MAE (percentage points; lower is better)}} \\
    $k$ & Exec. (\%) & Random & Weighted & gp-IRT \\
    \midrule
    10 & 1.9 & 12.09 & \underline{11.20} & \textbf{9.69} \\
    20 & 3.9 & 8.21 & \underline{7.20} & \textbf{7.01} \\
    30 & 5.8 & \textbf{6.58} & \underline{6.79} & 6.96 \\
    50 & 9.6 & \textbf{5.16} & 5.89 & \underline{5.65} \\
    75 & 14.5 & 4.03 & \underline{3.27} & \textbf{3.21} \\
    100 & 19.3 & \textbf{3.32} & \underline{3.35} & 3.55 \\
    150 & 28.9 & \underline{2.47} & 2.54 & \textbf{2.43} \\
    200 & 38.5 & 2.05 & \underline{1.88} & \textbf{1.67} \\
    250 & 48.2 & 1.72 & \underline{1.24} & \textbf{1.18} \\
    300 & 57.8 & 1.42 & \underline{1.38} & \textbf{1.34} \\
    350 & 67.4 & \underline{1.16} & 1.25 & \textbf{1.15} \\
    400 & 77.1 & \textbf{0.94} & 1.34 & \underline{1.13} \\
    \midrule
    \multicolumn{5}{l}{\textit{(b) Spearman correlation (higher is better)}} \\
    $k$ & Exec. (\%) & Random & Weighted & gp-IRT \\
    \midrule
    10 & 1.9 & 0.539 & \underline{0.634} & \textbf{0.638} \\
    20 & 3.9 & \textbf{0.664} & \underline{0.639} & 0.636 \\
    30 & 5.8 & \textbf{0.744} & \underline{0.625} & 0.620 \\
    50 & 9.6 & \textbf{0.809} & 0.758 & \underline{0.762} \\
    75 & 14.5 & \underline{0.853} & \textbf{0.861} & \textbf{0.861} \\
    100 & 19.3 & \textbf{0.886} & \underline{0.812} & 0.811 \\
    150 & 28.9 & \underline{0.924} & \textbf{0.933} & \textbf{0.933} \\
    200 & 38.5 & 0.940 & \underline{0.976} & \textbf{0.977} \\
    250 & 48.2 & 0.956 & \underline{0.975} & \textbf{0.976} \\
    300 & 57.8 & \textbf{0.964} & 0.961 & \underline{0.962} \\
    350 & 67.4 & \textbf{0.973} & \underline{0.969} & 0.967 \\
    400 & 77.1 & \textbf{0.980} & 0.967 & \underline{0.970} \\
    \midrule
    \multicolumn{5}{l}{\textit{(c) Kendall's Tau (higher is better)}} \\
    $k$ & Exec. (\%) & Random & Weighted & gp-IRT \\
    \midrule
    10 & 1.9 & 0.414 & \underline{0.499} & \textbf{0.502} \\
    20 & 3.9 & \textbf{0.510} & \underline{0.486} & 0.481 \\
    30 & 5.8 & \textbf{0.580} & \underline{0.469} & 0.464 \\
    50 & 9.6 & \textbf{0.649} & 0.581 & \underline{0.583} \\
    75 & 14.5 & \underline{0.697} & \textbf{0.735} & \textbf{0.735} \\
    100 & 19.3 & \textbf{0.736} & \underline{0.668} & 0.667 \\
    150 & 28.9 & 0.786 & \textbf{0.811} & \underline{0.809} \\
    200 & 38.5 & 0.820 & \textbf{0.894} & \underline{0.893} \\
    250 & 48.2 & 0.848 & \underline{0.883} & \textbf{0.884} \\
    300 & 57.8 & \underline{0.865} & 0.862 & \textbf{0.866} \\
    350 & 67.4 & \textbf{0.889} & \underline{0.884} & 0.876 \\
    400 & 77.1 & \textbf{0.909} & 0.868 & \underline{0.876} \\
    \bottomrule
  \end{tabular}
  \caption{Complete transfer results for anonymized Family D. Bold and underlining mark the best and second-best displayed values at each budget, respectively.}
  \label{tab:transfer-family-d-complete}
\end{table*}

%% file: tables/appendix_transfer_family_e.tex
\begin{table*}[t]
  \centering
  \small
  \setlength{\tabcolsep}{3.5pt}
  \begin{tabular}{rrrrr}
    \toprule
    \multicolumn{5}{l}{\textit{(a) Pass-rate MAE (percentage points; lower is better)}} \\
    $k$ & Exec. (\%) & Random & Weighted & gp-IRT \\
    \midrule
    10 & 1.9 & \underline{9.88} & 10.61 & \textbf{8.38} \\
    20 & 3.9 & 6.72 & \underline{4.79} & \textbf{4.09} \\
    30 & 5.8 & 5.80 & \underline{5.33} & \textbf{4.40} \\
    50 & 9.6 & \textbf{4.20} & 6.09 & \underline{5.54} \\
    75 & 14.5 & \underline{3.30} & 3.32 & \textbf{3.05} \\
    100 & 19.3 & 2.70 & \textbf{1.55} & \underline{1.61} \\
    150 & 28.9 & 2.01 & \underline{1.69} & \textbf{1.60} \\
    200 & 38.5 & \textbf{1.71} & 1.89 & \underline{1.74} \\
    250 & 48.2 & \textbf{1.40} & 1.60 & \underline{1.59} \\
    300 & 57.8 & 1.02 & \textbf{0.84} & \underline{0.94} \\
    350 & 67.4 & \textbf{0.81} & \underline{0.94} & 1.03 \\
    400 & 77.1 & \textbf{0.62} & 0.66 & \underline{0.65} \\
    \midrule
    \multicolumn{5}{l}{\textit{(b) Spearman correlation (higher is better)}} \\
    $k$ & Exec. (\%) & Random & Weighted & gp-IRT \\
    \midrule
    10 & 1.9 & 0.464 & \underline{0.810} & \textbf{0.837} \\
    20 & 3.9 & \textbf{0.646} & 0.624 & \underline{0.629} \\
    30 & 5.8 & 0.684 & \underline{0.765} & \textbf{0.797} \\
    50 & 9.6 & 0.740 & \underline{0.846} & \textbf{0.888} \\
    75 & 14.5 & \underline{0.806} & \textbf{0.823} & \textbf{0.823} \\
    100 & 19.3 & 0.847 & \underline{0.951} & \textbf{0.986} \\
    150 & 28.9 & \underline{0.873} & 0.837 & \textbf{0.881} \\
    200 & 38.5 & \textbf{0.894} & \underline{0.881} & \underline{0.881} \\
    250 & 48.2 & \underline{0.898} & \textbf{0.923} & \textbf{0.923} \\
    300 & 57.8 & \underline{0.914} & \textbf{0.958} & \textbf{0.958} \\
    350 & 67.4 & \underline{0.933} & \textbf{0.972} & \textbf{0.972} \\
    400 & 77.1 & \underline{0.960} & \textbf{0.993} & \textbf{0.993} \\
    \midrule
    \multicolumn{5}{l}{\textit{(c) Kendall's Tau (higher is better)}} \\
    $k$ & Exec. (\%) & Random & Weighted & gp-IRT \\
    \midrule
    10 & 1.9 & 0.364 & \underline{0.647} & \textbf{0.657} \\
    20 & 3.9 & \textbf{0.522} & \underline{0.465} & 0.455 \\
    30 & 5.8 & 0.551 & \underline{0.646} & \textbf{0.667} \\
    50 & 9.6 & 0.600 & \underline{0.697} & \textbf{0.727} \\
    75 & 14.5 & \textbf{0.669} & \underline{0.657} & \underline{0.657} \\
    100 & 19.3 & 0.716 & \underline{0.879} & \textbf{0.939} \\
    150 & 28.9 & \underline{0.749} & 0.718 & \textbf{0.758} \\
    200 & 38.5 & \textbf{0.785} & \underline{0.758} & \underline{0.758} \\
    250 & 48.2 & \textbf{0.792} & \underline{0.788} & \underline{0.788} \\
    300 & 57.8 & \underline{0.822} & \textbf{0.848} & \textbf{0.848} \\
    350 & 67.4 & \underline{0.852} & \textbf{0.909} & \textbf{0.909} \\
    400 & 77.1 & \underline{0.893} & \textbf{0.970} & \textbf{0.970} \\
    \bottomrule
  \end{tabular}
  \caption{Complete transfer results for anonymized Family E. Bold and underlining mark the best and second-best displayed values at each budget, respectively.}
  \label{tab:transfer-family-e-complete}
\end{table*}

%% file: tables/appendix_calibration_windows.tex
\begin{table*}[t]
  \centering
  \scriptsize
  \setlength{\tabcolsep}{2.8pt}
  \begin{tabular}{lrrrrrrrrr}
    \toprule
    \multicolumn{10}{l}{\textit{(a) Pass-rate MAE (percentage points; lower is better)}} \\
    \multicolumn{2}{c}{} & \multicolumn{2}{c}{${k=100}$} & \multicolumn{2}{c}{${k=200}$} & \multicolumn{2}{c}{${k=300}$} & \multicolumn{2}{c}{${k=400}$} \\
    Start & Runs & Weighted & gp-IRT & Weighted & gp-IRT & Weighted & gp-IRT & Weighted & gp-IRT \\
    \midrule
    Day 1 & 287 & 2.71 & 2.65 & 1.43 & 1.40 & 1.06 & 0.99 & 0.51 & 0.47 \\
    Day 4 & 258 & 2.86 & 2.97 & 1.42 & 1.35 & 1.14 & 1.06 & 0.53 & 0.52 \\
    Day 7 & 222 & 2.71 & 2.69 & 1.50 & 1.49 & 1.29 & 1.35 & 0.50 & 0.45 \\
    Day 10 & 192 & 2.39 & 2.36 & 1.66 & 1.59 & 0.97 & 0.98 & 0.52 & 0.52 \\
    Day 13 & 157 & 2.12 & 2.04 & 1.43 & 1.35 & 1.16 & 1.05 & 0.55 & 0.52 \\
    Day 16 & 128 & 2.33 & 2.30 & 1.49 & 1.45 & 0.95 & 0.93 & 0.51 & 0.50 \\
    Day 19 & 101 & 2.14 & 2.15 & 1.59 & 1.59 & 0.97 & 0.94 & 0.49 & 0.43 \\
    Day 22 & 74 & 2.06 & 2.04 & 1.46 & 1.49 & 1.12 & 1.14 & 0.58 & 0.56 \\
    Day 25 & 43 & 2.54 & 2.46 & 1.60 & 1.58 & 1.22 & 1.14 & 0.58 & 0.52 \\
    Day 28 & 14 & 2.20 & 2.19 & 1.45 & 1.41 & 1.04 & 1.03 & 0.51 & 0.51 \\
    \midrule
    \multicolumn{10}{l}{\textit{(b) Spearman correlation (higher is better)}} \\
    \multicolumn{2}{c}{} & \multicolumn{2}{c}{${k=100}$} & \multicolumn{2}{c}{${k=200}$} & \multicolumn{2}{c}{${k=300}$} & \multicolumn{2}{c}{${k=400}$} \\
    Start & Runs & Weighted & gp-IRT & Weighted & gp-IRT & Weighted & gp-IRT & Weighted & gp-IRT \\
    \midrule
    Day 1 & 287 & 0.930 & 0.930 & 0.972 & 0.972 & 0.980 & 0.981 & 0.993 & 0.994 \\
    Day 4 & 258 & 0.903 & 0.903 & 0.980 & 0.980 & 0.981 & 0.982 & 0.994 & 0.995 \\
    Day 7 & 222 & 0.920 & 0.923 & 0.967 & 0.968 & 0.978 & 0.978 & 0.994 & 0.995 \\
    Day 10 & 192 & 0.934 & 0.934 & 0.968 & 0.969 & 0.984 & 0.984 & 0.995 & 0.995 \\
    Day 13 & 157 & 0.948 & 0.951 & 0.974 & 0.975 & 0.981 & 0.983 & 0.994 & 0.994 \\
    Day 16 & 128 & 0.947 & 0.947 & 0.973 & 0.973 & 0.983 & 0.984 & 0.994 & 0.994 \\
    Day 19 & 101 & 0.952 & 0.953 & 0.972 & 0.972 & 0.981 & 0.982 & 0.995 & 0.996 \\
    Day 22 & 74 & 0.952 & 0.955 & 0.969 & 0.969 & 0.981 & 0.981 & 0.994 & 0.994 \\
    Day 25 & 43 & 0.925 & 0.925 & 0.967 & 0.968 & 0.984 & 0.985 & 0.995 & 0.995 \\
    Day 28 & 14 & 0.947 & 0.948 & 0.977 & 0.977 & 0.980 & 0.980 & 0.995 & 0.995 \\
    \midrule
    \multicolumn{10}{l}{\textit{(c) Kendall's Tau (higher is better)}} \\
    \multicolumn{2}{c}{} & \multicolumn{2}{c}{${k=100}$} & \multicolumn{2}{c}{${k=200}$} & \multicolumn{2}{c}{${k=300}$} & \multicolumn{2}{c}{${k=400}$} \\
    Start & Runs & Weighted & gp-IRT & Weighted & gp-IRT & Weighted & gp-IRT & Weighted & gp-IRT \\
    \midrule
    Day 1 & 287 & 0.771 & 0.770 & 0.857 & 0.859 & 0.884 & 0.887 & 0.937 & 0.941 \\
    Day 4 & 258 & 0.740 & 0.739 & 0.883 & 0.883 & 0.889 & 0.894 & 0.942 & 0.945 \\
    Day 7 & 222 & 0.760 & 0.762 & 0.854 & 0.854 & 0.879 & 0.877 & 0.944 & 0.948 \\
    Day 10 & 192 & 0.779 & 0.777 & 0.851 & 0.854 & 0.899 & 0.898 & 0.945 & 0.947 \\
    Day 13 & 157 & 0.803 & 0.809 & 0.865 & 0.868 & 0.888 & 0.893 & 0.939 & 0.944 \\
    Day 16 & 128 & 0.798 & 0.799 & 0.860 & 0.861 & 0.895 & 0.897 & 0.943 & 0.947 \\
    Day 19 & 101 & 0.812 & 0.814 & 0.858 & 0.858 & 0.889 & 0.892 & 0.948 & 0.952 \\
    Day 22 & 74 & 0.811 & 0.814 & 0.851 & 0.852 & 0.883 & 0.885 & 0.941 & 0.943 \\
    Day 25 & 43 & 0.756 & 0.757 & 0.847 & 0.849 & 0.897 & 0.901 & 0.945 & 0.949 \\
    Day 28 & 14 & 0.799 & 0.801 & 0.871 & 0.872 & 0.887 & 0.890 & 0.945 & 0.947 \\
    \bottomrule
  \end{tabular}
  \caption{Complete calibration-window sensitivity results. All windows end Day 28 and use the same 287 held-out runs. MAE is in percentage points. Values are not emphasized because the windows form a sensitivity sweep rather than prespecified competing methods.}
  \label{tab:calibration-window-complete}
\end{table*}